\documentclass[times,twocolumn,final,authoryear]{elsarticle_arxiv}
\usepackage[a4paper, total={6.8in, 8in}]{geometry}

\usepackage{ycviu}
\usepackage{framed,multirow}

\usepackage{amssymb}
\usepackage{latexsym}

\usepackage{url}
\usepackage{xcolor}
\definecolor{newcolor}{rgb}{.8,.349,.1}
\usepackage{balance}
\usepackage{times}
\usepackage{epsfig}
\usepackage{graphicx}
\usepackage{amsmath}
\usepackage{amssymb}
\usepackage{colortbl}
\usepackage{multirow}
\usepackage{caption}
\usepackage{subcaption}
\usepackage{multibib}
\newcites{latex}{References}
\usepackage{tcolorbox}
\usepackage{enumitem}
\usepackage{xspace}
\usepackage{dsfont}

\definecolor{LightCyan}{rgb}{0.88,1,0.88}
\definecolor{LightCyan2}{rgb}{0.0,0.7,0.0}
\definecolor{DarkGreen}{rgb}{0.0,0.0,0.7}
\definecolor{bananamania}{rgb}{0.98, 0.91, 0.71}

\newcommand{\mycomment}[1]{}

\newcommand{\mIdent}{\boldsymbol{\mathds{I}}}
\newcommand{\tabincell}[2]{\begin{tabular}{@{}#1@{}}#2\end{tabular}}

\makeatletter
\DeclareRobustCommand\onedot{\futurelet\@let@token\bmv@onedotaux}
\def\bmv@onedotaux{\ifx\@let@token.\else.\null\fi\xspace}
\def\eg{\emph{e.g}\onedot} 
\def\ie{\emph{i.e}\onedot} 
 
\def\etc{\emph{etc}\onedot} 
\def\wrt{w.r.t\onedot}

\makeatother

\begin{document}

\begin{frontmatter}
\title{Multivariate Prototype Representation for Domain-Generalized Incremental Learning\vspace{-0.5cm}}

\author[1]{Can \snm{Peng}}
\author[1,3]{Piotr \snm{Koniusz}\corref{cor1}}
\cortext[cor1]{Corresponding author. 
  }
\author[2]{Kaiyu \snm{Guo}}
\author[2]{Brian C. \snm{Lovell}}
\author[1,4]{Peyman \snm{Moghadam}\vspace{-0.2cm}}

\address[1]{Data61, CSIRO, Brisbane, Queensland, Australia}
\address[2]{The University of Queensland, Brisbane, Queensland, Australia}
\address[3]{The Australian National University, Canberra, Autralia}
\address[4]{Queensland University of Technology, Brisbane, Queensland, Australia}

\begin{abstract}
Deep learning models suffer from catastrophic forgetting when being fine-tuned with samples of new classes. This issue becomes even more pronounced when faced with the domain shift between training and testing data.
In this paper, we study the critical and less explored Domain-Generalized Class-Incremental Learning (DGCIL).
We design a DGCIL approach that remembers old classes, adapts to new classes, and can classify reliably objects from unseen domains.
Specifically, our loss formulation maintains classification boundaries and suppresses the domain-specific information of each class.
With no old exemplars stored, we use knowledge distillation and estimate old class prototype drift as incremental training advances. 
Our prototype representations are based on multivariate Normal distributions whose means and covariances  are constantly adapted to changing model features to represent old classes well by adapting to the feature space drift. 
For old classes, we sample pseudo-features from the adapted Normal distributions with the help of Cholesky decomposition.
In contrast to previous pseudo-feature sampling strategies that rely solely on average mean prototypes, our method excels at capturing varying semantic information. 
Experiments on several benchmarks validate our claims. 

\end{abstract}
\end{frontmatter}


\vspace{-0.5cm}
\section{Introduction}
\label{sec: Introduction}
Despite the progress in deep learning, many algorithms fail when the data distribution changes. 
Objects of interest from new classes can continually appear, and simultaneously the old class data might be inaccessible due to data storage limitations, privacy, or licensing issues. 
Under such conditions, direct fine-tuning of deep learning models with the new class data will make a deep learning model  lose the performance on previously-learnt tasks -- the catastrophic forgetting problem. 
Also, when trained models are deployed into new data domains stemming from diverse sources of data, the model often fails to generalize to out-of-distribution cases. 
Incremental Learning (IL) addresses the problem of catastrophic forgetting posed by the incoming data whose distribution changes over time.
Domain Generalization (DG) tackles the out-of-distribution problem caused by the distribution shift between training and testing data. 
Unfortunately, IL and DG are usually considered separate problems.
\begin{figure}[t]
  \hspace{-0.3cm}
   \includegraphics[width=1.05\linewidth]{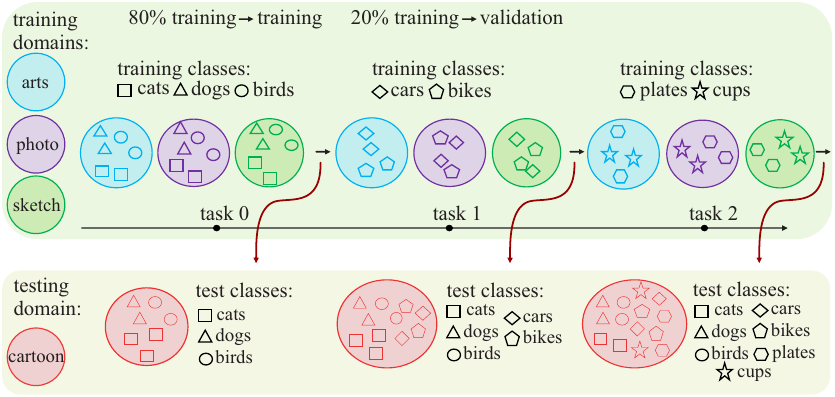}
   \vspace{-0.6cm}
    \caption{Domain-Generalized Class-Incremental Learning.
    Samples of new classes are provided from a fixed set of training domains for each incremental learning step $t$.  
    During testing, the model has to perform well on all the categories learnt so far under an unseen testing domain. 
    Each domain becomes a testing domain once, \ie, four separate runs are needed, over which average accuracy is calculated.}
	\label{fig1: DGCIL flowchart}
	\vspace{-0.3cm}
\end{figure}
Yet, for autonomous vehicles or robots, new classes of objects such as new road signs, new means of transport, and various surrounding buildings can continually appear and need to be learnt without forgetting the already-learnt visual concepts. 
The system also has to perform well under conditions that differ from the training domain, including variations in lighting, weather, infrastructure designs, \etc 
Domain-Generalized Class-Incremental Learning (DGCIL) addresses these two scenarios that require simultaneous handling of catastrophic forgetting and out-of-distribution recognition.
DGCIL approaches have to achieve good performance under novel test domains and a significant intake of new visual categories.

\begin{figure}[tbp]
\hspace{-0.30cm}
	\includegraphics[width=1.05\linewidth]{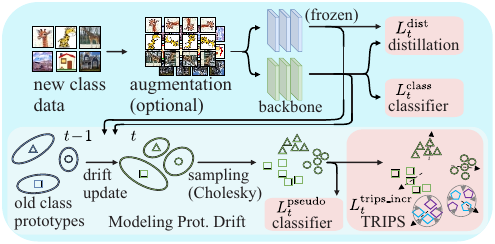}
	\vspace{-0.6cm}
    \caption{The TRIPS pipeline. 
    During the incremental step $t$, we pass new class samples through the old (frozen) and the current models.
    A classifier loss $L_t^\text{class}$ and a distillation loss $L_t^\text{dist}$ are applied. 
    The mean and covariance matrices of each class prototype are updated due to the feature drift between models. 
    Subsequently, pseudo-samples are obtained from covariances and passed to pseudo-classifier $L_t^\text{pseudo}$. 
    Finally, our TRIPS pushes feature vectors of different classes apart (black dashed arrows), while suppressing domain-specific information (grey arrows facing inwards).}
	\label{fig2: pipeline}
\end{figure}

Figure \ref{fig1: DGCIL flowchart} illustrates the DGCIL paradigm.   
During training, for each incremental step, with only new class data available, the model needs to learn representations of new classes as well as maintain representations of old classes.   
At the inference stage, the model is required to perform well in all the classes learnt so far on novel domains (not seen in training). 
Compared with IL and DG, DGCIL is more challenging and less explored. 
While some efforts have been made to explore continual domain adaptation which evaluates the model on continually arriving classes and domains \citep{buzzega2020dark, knights2022incloud, volpi2021continual, xie2022general, liu2023unsupervised}, 
DGCIL simultaneously  targets out-of-domain generalization. 
In contrast, the problem of continual domain adaptation focuses on adaptation under training domain shift. 
To the best of our knowledge,  MSL\_MOV \citep{simon2022generalizing} is the only work that targets a similar task. 
However, in their setting, old class exemplars from each training domain need to be stored which might cause data privacy or storage issues, \ie, storing exemplars may be simply not an option. In contrast, in this work, we design  an exemplar-free DGCIL model. 

As we are interested in an exemplar-free setting, we remove the exemplars from the experimental setting. For a baseline, we modify the only existing DGCIL model, MSL\_MOV, to work with the prototypes instead of exemplars. 

Moreover, we notice that the evaluation protocol introduced by MSL\_MOV is limited to their specific pipeline design, and may be biased towards old classes.
Specifically, within each incremental step, MSL\_MOV performs three-stage training (training for each domain progressively). For each incremental step, MSL\_MOV reports the average results from
three stages. The evaluation results within such a training procedure may be  not universal as they are biased towards first- and early-stage results, \eg, the first stage of a new incremental step may be biased towards good performance on old classes, whereas the last stage may be biased towards new classes. Also, the first stage may be more aligned with the test domain than the model from all three stages. Thus, averaging  the results of three stages may make results on old classes look better  compared to a one-off evaluation at the end of the incremental step.

Thus, we propose a more realistic evaluation protocol as follows. During each
incremental step, we only perform a one-time evaluation once the given incremental step learnt from all training domains (all stages are completed). Based on the publicly available domain generalization benchmark DomainBed \citep{gulrajani2020search}, we build our framework and method. We reproduce MSL\_MOV in our evaluation setting for fair comparisons with our approach. Figure \ref{fig1: DGCIL_eval_prot} illustrates the difference in both evaluation settings.

As we adopt the challenging exemplar-free DGCIL setting, we propose {\em TRIplet loss with Pseudo old-class feature Sampling} (TRIPS). 
Figure \ref{fig2: pipeline} illustrates our model.  
During training, we adopt a novel variant of triplet loss that focuses on distinguishing semantic information while suppressing the domain information. The goal of such a loss is to improve the generalization of the model towards an unseen test distribution.
In addition, we apply knowledge distillation to maintain previously-learnt knowledge and prevent catastrophic forgetting. 
Moreover, old classes are described by class prototype representations based on the multivariate Normal distributions.  
As incremental sessions progress, the drift associated with each mean and covariance matrix of each prototype is estimated, and prototype representations are updated accordingly before we sample old-class pseudo-features via the Cholesky decomposition. 
Such pseudo-features help preserve the knowledge of old classes from the past sessions and maintain the class separation by TRIPS, facilitating balanced performance on old and new class concepts. 

\vspace{0.2cm}
Our main contributions are summarised as follows:
\vspace{0.1cm}
\renewcommand{\labelenumi}{\roman{enumi}.}
\vspace{-0.2cm}
\begin{enumerate}[leftmargin=0.5cm]
\item  With data storage, privacy, and licensing issues in mind, we propose an exemplar-free approach for  DGCIL, based on {\em TRIplet loss with Pseudo
old-class feature Sampling}. The goal of TRIPS is to keep features of different classes apart while suppressing domain-related feature information, which facilitates domain generalization.
\vspace{-0.2cm}
\item As TRIPS has to handle old and new classes incrementally, we introduce class-wise prototype representations based on the multivariate Normal distributions. We estimate their drift (for each mean/covariance) as the backbone is updated along incremental sessions. We then sample pseudo-features for old classes from  such distributions. 
\vspace{-0.2cm}
\item  Details of model selection (validation procedure), data splitting, and evaluation strategies are important for reproducibility. %
As  exemplar-free DGCIL models do not exist, we provide a comprehensive DGCIL task setting for exemplar-free DGCIL including 
(i) validation procedure,
(ii) evaluation setting reflecting the performance of each incremental step, and
(iii) the overall and balanced performance based on the domain-wise and class-wise average and harmonic accuracy, respectively. 
The code with validation/evaluation protocols will be released.
\end{enumerate}

\section{Related Works}
\label{sec: Related Works}

In this section, we explain related works in Incremental Learning (IL), Domain Generalization (DG), and Domain-Generalized Class-IL (DGCIL).

\vspace{0.1cm}
\noindent\textbf{Incremental Learning Settings.} 
IL requires the model to continually learn new tasks from a sequential data stream without forgetting the past.  
IL can be categorized into three sub-tasks: Task-IL, Class-IL, and General IL. 
The main difference between them is whether task boundaries are accessible during training and testing. 
Task-IL, the simplest IL scenario, allows access to task boundaries during both training and testing \citep{li2017learning}.
Different from Task-IL, the Class-IL scenario only provides task boundaries during training \citep{kirkpatrick2017overcoming, chaudhry2018riemannian, aljundi2018memory, rebuffi2017icarl, hou2019learning}. 
During testing, a Class-IL model is required to have a unified classifier that can classify all classes learnt so far. 
General IL, a newly emerged IL scenario, is a more practical setting than Task-IL and Class-IL. 
In this setting, the task boundaries are not accessible during either training or testing \citep{xie2022general, ji2022complementary, li2022learning}. 
However, General-IL does not model out-of-distribution scenes to facilitate novel test distributions.
Some General-IL methods consider domain adaptation \citep{buzzega2020dark, xie2022general} by incrementally adding new classes/domains and performing testing on classes/domains learnt so far. 
They do not consider unseen domains during testing.
In contrast, we study Domain-Generalized IL to preserve old class knowledge, and continually learn new class concepts, while generalizing well to unseen test distributions. 

\vspace{0.1cm}
\noindent\textbf{Solutions to Incremental Learning.} 
Current IL methods are often based on: regularization, dynamic architecture, or old exemplar replay.
Advanced regularization terms impose constraints on the network update and mitigate catastrophic forgetting. 
Some models regularize the parameter updating according to their importance \wrt previously-learnt tasks \citep{kirkpatrick2017overcoming, chaudhry2018riemannian, aljundi2018memory}. 
Other models maintain past feature distribution by knowledge distillation on logits~\citep{li2017learning, rebuffi2017icarl, hou2019learning} or on intermediate features~\citep{simon2021learning, roy2023subspace}.
Dynamic architecture \citep{yoon2018lifelong, douillard2022dytox} assigns each new task with a task-specific sub-network to balance the stability and plasticity trade-off.
Old exemplar replay retains some old class samples for experience reply during  new task training to refresh the model memory \citep{rebuffi2017icarl, hou2019learning, xie2022general, li2022learning, ji2022complementary}. 

\vspace{0.1cm}
\noindent\textbf{Domain Generalization.} 
DG approaches are required to make good predictions on novel test domains. 
Many DG models reduce domain gaps in the latent space and obtain easy-to-transfer model parameters by meta-learning \citep{dou2019domain}, data augmentation \citep{zhang2017mixup, zhou2021domain}, or capturing causal relations \citep{arjovsky2019invariant, krueger2021out}. 
Although a myriad of domain generalization algorithms has been proposed, they use various experimental settings, such as different datasets and model selection criteria \citep{gulrajani2020search}. 
\cite{gulrajani2020search} propose a benchmark testbed called DomainBed and argue that when equipped with modern neural network architectures and data augmentation techniques, empirical risk minimization (ERM) achieves state-of-the-art performance.

\vspace{0.1cm}
\noindent\textbf{Domain-Generalized Class-IL (DGCIL).} 
In this paper, we study Domain-Generalized Class-Incremental Learning, bridging the gap between the Class-IL (CIL) and DG problems. 
DGCIL solves both catastrophic forgetting due to the semantic shift and domain generalization beyond training domains.  
Compared with CIL and DG, DGCIL is more challenging and less explored. 
 MSL\_MOV \citep{simon2022generalizing} is the only related work that explores a similar problem by the use of Mahalanobis metrics for classification in unseen domains. 
MSL\_MOV also uses old class exemplars and exponential moving average knowledge distillation to overcome catastrophic forgetting. 
However, in many situations, old class exemplars should not be used (privacy, storage, \etc) and the number of parameters of MSL\_MOV grows linearly. 
In this paper, we provide a comprehensive task setting for DGCIL and propose an exemplar-free DGCIL method.

\section{What is the DGCIL About?}
\label{sec: Problem Formulation}

A DGCIL model (similar to CIL) learns new tasks continually in several steps but the model is required to perform well on unseen test domains (domain-invariant model).
The DGCIL algorithms have two goals: 
(i) learning new class concepts while maintaining knowledge of old class concepts, 
and (ii) learning semantic invariance from training data with domain-specific information.
The latter property helps the model generalize to out-of-distribution cases during testing. 
Figure \ref{fig1: DGCIL flowchart} describes the training and testing of DGCIL.

Specifically, a DGCIL task has $T+1$ steps each associated with a sub-task $\mathcal{T}_t$ where $t\in\{0, 1, \ldots,T\}$. 
Let $\mathcal{T}_0$ denote the base task and $\{\mathcal{T}_t\!:\! t\geq 1\}$ be  incremental tasks. 
The $t$-th task has to learn $\lvert\mathcal{C}_t\rvert$  new class concepts with training samples of   categories from set $\mathcal{C}_t$.
Considering the possible memory limitation in realistic conditions, we opt for an exemplar-free setting. 
Training data from different tasks has no overlap in class labels or samples, thus $\mathcal{C}_t \cap \mathcal{C}_{t'} = \varnothing$ if $t\neq t'$.
During each incremental learning task, training samples from $Z$ different domain distributions are provided by sets $\mathcal{D}_z$  where $z \in \{1, \ldots, Z\}$.  
The total training data is represented as $\mathcal{D}_\text{train}=\mathcal{D}_1\cup\mathcal{D}_2 \ldots \cup \mathcal{D}_Z$.
During testing, the model will be evaluated on all classes learnt up to $t$ in an unseen domain $\mathcal{D}_\text{test}=\mathcal{D}_{Z+1}$.
Thus, $\mathcal{C}_\text{train}=\mathcal{C}_t$ and $\mathcal{C}_\text{test}= \mathcal{C}_0\cup\mathcal{C}_1\ldots\cup \mathcal{C}_t$, while domains
$\mathcal{Z}_\text{train}\cap \mathcal{Z}_\text{test}=\varnothing$.

\section{Proposed Approach}
\label{sec: Algorithm}
In this section, we outline our  algorithm, which consists of:
(i) triplet loss which teaches the model to distinguish semantic information across different classes while suppressing domain-specific information, 
(ii) cross-entropy knowledge distillation which prevents catastrophic forgetting of past knowledge, 
(iii) prototype representations based on the multivariate Normal distribution with a drift mechanism that adapts prototypes from previous sessions to the feature space of the current session, 
and (iv) sampling mechanism via Cholesky decomposition.  

\subsection{Learning New Classes}
\label{sec:lnc}
Our approach consists of three components: (i) learning new knowledge, (ii) preserving the former knowledge, and (iii) handling out-of-distribution cases. 

For the first component, we adopt the commonly used strategy based on the cross-entropy loss using the ground truth labels and softmax-normalized model output over all classes learnt so far.
Let a set of sample-label pairs $(\mathbf{x},y)$ for the current session $t$ be stored in batch $\mathcal{B}$. 
Let the set of classes observed so far for the session/step $t$ where $t\leq T$ be denoted as $\mathcal{C}'=\mathcal{C}_0\cup\mathcal{C}_1\ldots\cup\mathcal{C}_t$. Then we have the new knowledge learning for the step $t$:
\begin{equation}
    L_{t}^\text{class} = - \frac{1}{\lvert\mathcal{B}\lvert} \sum\limits_{(\mathbf{x},y) \in \mathcal{B}} \!\!\log\!\Bigg(\frac{\exp\big(\boldsymbol{\theta}_{ty}^\top f_t(\mathbf{x})\big)}{\sum\limits_{c\in \mathcal{C}'} \exp\big(\boldsymbol{\theta}_{tc}^\top f_t(\mathbf{x})\big)}
    \Bigg).
	\label{eq: cross_entropy_loss}
\end{equation}
\noindent 
Notice that while $\mathcal{B}$ only contains samples of classes in set $\mathcal{C}_t$, the denominator of Eq. \eqref{eq: cross_entropy_loss} runs over all classes observed in sessions/steps $0,\ldots,t$. 
Function $f_t(\cdot)$ denotes our feature extractor trained in session $t$. Parameters of $f_t(\cdot)$ are initialized by coping from parameters of $f_{t-1}(\cdot)$. Moreover, $\boldsymbol{\theta}_{ty}$ denotes a linear projection from session $t$ for class $y\in\mathcal{C}'$ (which is essentially an FC layer). 
Linear projection parameters $\boldsymbol{\theta}_{ty'}$ for $y'\in \mathcal{C}''$ where $\mathcal{C}''\!=\mathcal{C}'\setminus\mathcal{C}_t$ are initialized by coping from $\boldsymbol{\theta}_{t'y'}$ where $t'=t-1$. In practice, we also use bias terms $b_{ty}$, \ie, we have a linear projection $\boldsymbol{\theta}_{ty}^\top f_t(\mathbf{x})+b_{ty}$ but we skip $b_{ty}$ for brevity. 

\subsection{Knowledge Distillation}
\label{sec:kd}
As DGCIL approaches have to retain the knowledge from previous sessions and overcome catastrophic forgetting \citep{hinton2015distilling}, for our model we adopt simple knowledge distillation\footnote{
Kindly note we do not claim the methodology of Sections \ref{sec:lnc} and \ref{sec:kd} {\em per se} as contributions. 
These are prerequisites required in our model.} 
to transfer responses of feature extractor $f_{t-1}(\cdot)$ to $f_{t}(\cdot)$ and linear projection $\boldsymbol{\theta}_{t-1}$ to $\boldsymbol{\theta}_{t}$ by cross-entropy.
The samples of new classes are passed via frozen $f_{t-1}(\cdot)$ and $\boldsymbol{\theta}_{t-1}$ to distill through the lens of old classes from set $\mathcal{C}''\!=\mathcal{C}_0\cup\mathcal{C}_1\ldots\cup\mathcal{C}_{t-1}$ representing past sessions $t=0,\ldots,t-1$ into $f_{t}(\cdot)$ and $\boldsymbol{\theta}_{t}$. 
We define distillation for the step $t$ as:
\begin{align}
    &L_{t}^\text{dist} = - \frac{1}{\lvert\mathcal{B}\rvert} \sum\limits_{\mathbf{x} \in \mathcal{B}} \;\sum\limits_{c \in \mathcal{C}''} \pi_{c}^{t-1}(\mathbf{x}) \log\big(\pi_{c}^{t}(\mathbf{x})\big)
	\label{eq: distillation_loss}\\
	    &\quad\text{where}\quad\pi_{c}^t(\mathbf{x}) = \frac{\exp\big(\boldsymbol{\theta}_{tc}^\top f_t(\mathbf{x})\big)}{\sum\limits_{c'\in \mathcal{C}''} \exp\big(\boldsymbol{\theta}_{tc'}^\top f_t(\mathbf{x})\big)}.
	\label{eq: model_output}
\end{align}
\noindent In Eq. \eqref{eq: distillation_loss} and \eqref{eq: model_output}, $\boldsymbol{\pi}^{t}$ and $\boldsymbol{\pi}^{t-1}$ are the probability outputs of the current model ($t$) and the previous model ($t-1$) when passing samples of current session $t$, denoted as $(\mathbf{x},y)\in\mathcal{B}$, but distilling them through the lens of classes from the past sessions (set $\mathcal{C}''$). 
Although batch $\mathcal{B}$ contains pairs $(\mathbf{x},y)$, we simply enumerate over data samples and so we slightly abuse our notation by writing $\mathbf{x}\in\mathcal{B}$. 
We also use the bias terms but we skip them in equations for clarity.

\subsection{The TRIPS Loss}
As DGCIL approaches have to perform well during testing on an unseen domain (out-of-distribution conditions), during training, we want our model to capture the semantic information while suppressing the domain-specific information.
To this end, we adapt our TRIPS formulation based on the triplet loss\footnote{
Kindly note we do not claim standard triplet loss as our contribution. 
Our contribution is in the bespoke design that separates classes while discarding domain-specific information.}.

Specifically, we regard encoded feature vectors $f_t(\mathbf{x})$ and $f_t(\mathbf{x}')$ of samples $\mathbf{x}$ and $\mathbf{x}'$ with the same class labels $y(\mathbf{x})=y(\mathbf{x}')$ but different domain labels $z(\mathbf{x})\neq z(\mathbf{x}')$ as positive pairs, denoted as $\gamma^+_{\mathbf{x},\mathbf{x}'}$. 
Moreover, we regard feature vectors $f_t(\mathbf{x})$ and $f_t(\mathbf{x}')$  with two different class labels $y(\mathbf{x})\neq y(\mathbf{x}')$ but the same domain labels $z(\mathbf{x})=z(\mathbf{x}')$ as negative pairs, denoted as  $\gamma^-_{\mathbf{x},\mathbf{x}'}$.
To help the model learn to classify class concepts instead of domain concepts, the distance between positive pairs should be  smaller than the distance between negative pairs. 
We employ the squared Euclidean distance  $d(\mathbf{x},\mathbf{x}') = \|f_t(\mathbf{x}) - f_t(\mathbf{x}')\|_2^2$ to measure the pairwise distance. Formally, positive and negative feature pairs are defined as:
\begin{equation}
	\gamma^+_{\mathbf{x}.\mathbf{x}'} =
	\begin{cases}
	1\textrm{ }&\mbox{if} \textrm{ } y(\mathbf{x}) = y(\mathbf{x}') \textrm{ } \wedge \textrm{ } z(\mathbf{x}) \neq z(\mathbf{x}'), \\
	0\textrm{ }&\mbox{otherwise},
	\end{cases}
	\label{eq: positive_pair_matrix}
\end{equation}
\begin{equation}
	\delta^-_{\mathbf{x},\mathbf{x}'} =
	\begin{cases}
	1\textrm{ }&\mbox{if}\textrm{ }y(\mathbf{x}) \neq y(\mathbf{x}') \textrm{ } \wedge \textrm{ } z(\mathbf{x}) = z(\mathbf{x}'), \\
	0\textrm{ }&\mbox{otherwise}.
	\end{cases}
	\label{eq: negative_pair_matrix}
\end{equation}

\noindent Subsequently, we define our TRIPS loss as follows:
\begin{align}
& \!\!\!\!L_t^\text{trips\_base} = \frac{1}{\lvert\mathcal{B}\rvert} \sum\limits_{\mathbf{x}\in B}
\Big[\max_{\mathbf{x}'\in\mathcal{B}\setminus\{\mathbf{x}\}}\gamma^+_{\mathbf{x},\mathbf{x}'}\cdot d(\mathbf{x},\mathbf{x}')
\nonumber\\
    &\!\!\qquad\qquad\qquad\quad\!\Big.-\!\!\! \min_{\mathbf{x}''\in\mathcal{B}\setminus\{\mathbf{x}\}}\gamma^-_{\mathbf{x},\mathbf{x}''}\cdot d(\mathbf{x},\mathbf{x}'') + m\;\Big]_+,
	\label{eq: triplet_loss}
\end{align}
\noindent where $[\phi]_+\!\equiv\!\max(\phi,0)$ is a ReLU to realize the triplet loss with margin $m$ (set to $0$ in our experiments).  
Although batch $\mathcal{B}$ contains pairs $(\mathbf{x},y)$, we enumerate only over data samples--we slightly abuse our notation by writing $\mathbf{x}\in\mathcal{B}$.

\begin{figure}[tbp]
\hspace{-0.30cm}
	\centering
	\includegraphics[trim={2cm 15cm 5cm 5.5cm},clip, width=0.8\linewidth]
    {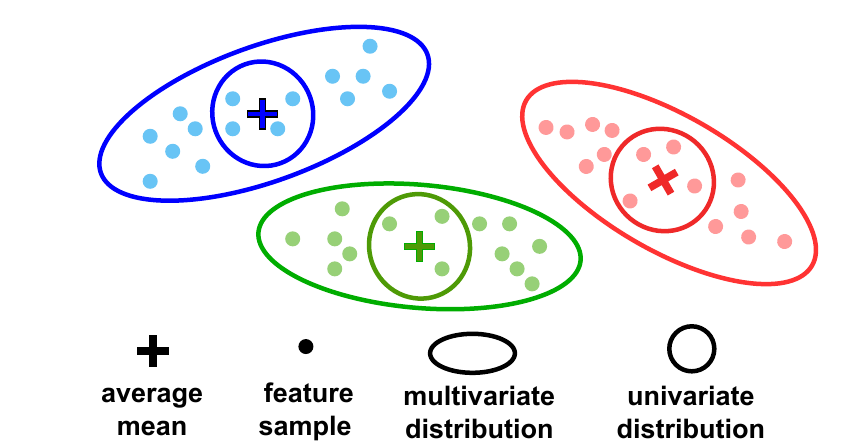}
	\vspace{-0.2cm}
    \caption{
    An example of the latent space and class-wise feature distributions. Due to the intra-class variance (multiple samples from several training domains), the single per-class feature mean (which is the first-order statistic) is not sufficient to represent the class distribution well. The old class distribution can be better captured by the mean and covariance matrix (the multivariate Normal distribution). The multivariate Normal distributions (second-order statistics) capture class boundaries better than isotropic/univariate Normal distributions and first-order statistics such as the mean vectors.
    }
	\label{fig3: covariance}
\end{figure}

\subsection{Modeling Prototype Drift}
\label{sec:pd}
With no access to old data, we utilize old-class prototype representations to prevent catastrophic forgetting. Related works on incremental learning tasks commonly use the per-class mean-based prototype \citep{yu2020semantic, zhu2021self}.
However, these prototypes are insufficient for capturing intra-class variance resulting from multiple samples from different training domains as they only capture first-order statistics of class-wise feature distribution.
In our approach, we utilize both the mean vector and covariance matrix for each class to enhance the representation of distribution of each old class. By incorporating covariance around its mean, the multivariate Normal distribution offers a more accurate model of class-specific boundaries. Furthermore, compared to isotropic/univariate Normal distributions or mean vectors (first-order statistics), considering second-order statistics through multivariate Normal distributions provides a more realistic distribution model.
Figure \ref{fig3: covariance} illustrates the  latent feature space with class-wise  distributions.

Given our prototype representations are modeled as the multivariate Normal distribution per class, we can draw pseudo-samples from these distributions for old classes. 
However, as the feature extractor changes over time, we need to account for the prototype drift in the evolving feature space. As the new classes are being learnt, the model is updated, and the feature distribution changes. 
Consequently, we cannot simply use the previously stored prototype representations for old classes without accounting for this drift.

With no access to old class exemplars, we capture the drift of features through the lens of samples of new classes passed through the current and past feature extractor, $f_t(\cdot)$ and $f_{t-1}(\cdot)$, respectively. Essentially, $f_t(\mathbf{x})-f_{t-1}(\mathbf{x})$ captures the changes. Such a modeling approach assumes that the semantics of old tasks and new tasks must somewhat overlap, \ie, the past feature space and current feature spaces are not completely disjoint semantically. We define:
\begin{align}
& \Delta\boldsymbol{\phi}(\mathbf{x})=f_t(\mathbf{x})-f_{t-1}(\mathbf{x}),\label{eq:delta_phi}\\
& w(\mathbf{x}, \boldsymbol{\mu})=\exp\Big(\!-\frac{1}{2\sigma^2}\lVert f_{t-1}(\mathbf{x})-\boldsymbol{\mu}\rVert_2^2\Big),\label{eq:weight}
\end{align}
where $\Delta\boldsymbol{\phi}(\mathbf{x})$ is the drift for an individual sample $\mathbf{x}$, and $w(\mathbf{x}, \boldsymbol{\mu}^{t-1}_c)$ captures similarity of $\mathbf{x}$ \wrt prototype mean $\boldsymbol{\mu}^{t-1}_c$ from step $t-1$ given some bandwidth hyper-parameter $\sigma>0$ (set to 0.5 in all our experiments). Let the drift of prototype mean be defined as:
\begin{align}
& \Delta\boldsymbol{\mu}_c=\frac{\sum_{\mathbf{x}\in\mathcal{B}}w(\mathbf{x}, \boldsymbol{\mu}^{t-1}_c)\cdot  \Delta\boldsymbol{\phi}(\mathbf{x})}{\sum_{\mathbf{x}'\in\mathcal{B}}w(\mathbf{x}', \boldsymbol{\mu}^{t-1}_c)},\\
& \Delta\boldsymbol{\mu}_c^b=\eta\Delta\boldsymbol{\mu}_c^{b-1}+(1-\eta)\Delta\boldsymbol{\mu}_c.\label{eq:exp1}
\end{align}
We go over batches $\mathcal{B}_b$ where $b=1,\ldots,B$ but we drop subscript $b$ from $\mathcal{B}_b$ for brevity. The final expression after $B$ batches is:
\begin{equation}
\boldsymbol{\mu}^t_c=\boldsymbol{\mu}^{t-1}_c+\Delta\boldsymbol{\mu}_c^{B}.
\end{equation}
Hyper-parameter $\eta$ (set to 0.1 in all our experiments) controls the so-called exponential moving average. 

Next, we estimate covariance drift ${\boldsymbol{\Sigma}'}_{\!\!c}$ for batches $\mathcal{B}_b$,  $b=1,\ldots,B$. We drop subscript $b$ from $\mathcal{B}_b$ for brevity.  As ${\boldsymbol{\Sigma}'}_{\!\!c}$ is estimated from a limited number of samples, we apply to it the shrinkage operator \citep{Ledoit110} known to estimate covariance matrices  reliably in such circumstances. 
We apply the exponential moving average to obtain ${\boldsymbol{\Sigma}'}_{\!\!c}^b$:
\begin{align}
& \!\!\!{\boldsymbol{\Sigma}'}_{\!\!c}=\frac{\sum\limits_{\mathbf{x}\in\mathcal{B}}w(\mathbf{x}, \boldsymbol{\mu}^{t-1}_c)\!\cdot\!  \big(f_t(\mathbf{x})\!-\!\boldsymbol{\mu}^t_c\big) \big(f_t(\mathbf{x})\!-\!\boldsymbol{\mu}^t_c\big)^\top }{\sum\limits_{\mathbf{x}'\in\mathcal{B}}w(\mathbf{x}', \boldsymbol{\mu}^{t-1}_c)},\label{eq:cov1}\\
& \!\!\!{\boldsymbol{\Sigma}'}_{\!\!c}^{b}=\eta{\boldsymbol{\Sigma}'}_{\!\!c}^{b-1}\!+\!(1-\eta)\big((1-\alpha){\boldsymbol{\Sigma}'}_{\!\!c}\!+\!\alpha\mIdent\big).\label{eq:cov2}
\end{align}
which leads to the final expression ${\boldsymbol{\Sigma}}_{c}^{t}\!=\!{\boldsymbol{\Sigma}'}_{\!\!c}^{B}$ after $B$ batches.
%
%
Hyper-parameter $\eta$ controls the exponential moving average (set to 0.1 in all our experiments). 
Moreover,  $0\leq\alpha\leq 1$ decides by how much covariance is shrunk towards the isotropic Normal distribution (set to 0.05 in all our experiments).

We then sample pseudo-features of past classes from our prototype representations during processing each  mini-batch $\mathcal{B}$.
\begin{equation}
\boldsymbol{\phi}^t_c \sim \mathcal{N}(\boldsymbol{\mu}_c^t, {\boldsymbol{\Sigma}}_{c}^{t})\quad\text{where}\quad c\sim\mathcal{U}(1, \lvert\mathcal{C}''\rvert),
\end{equation}
where $\mathcal{N}(\boldsymbol{\mu}_c^t, {\boldsymbol{\Sigma}}_{c}^{t})$ denotes the multivariate Normal distribution with mean $\boldsymbol{\mu}_c^t$ and covariance ${\boldsymbol{\Sigma}}_{c}^{t}$, and $\mathcal{U}(1, \lvert\mathcal{C}''\rvert)$ denotes the uniform distribution having $1/\lvert\mathcal{C}''\rvert$  probability within  support region $[1, \lvert\mathcal{C}''\rvert]$ (and zero elsewhere), and $\mathcal{C}''$ defined in Section \ref{sec:kd} is set of classes from sessions $0,\ldots,t-1$. 
Finally, we note that:
\begin{align}
&\boldsymbol{\phi}^t_c \sim \mathcal{N}(\boldsymbol{\mu}_c^t, {\boldsymbol{\Sigma}}_{c}^{t}) \quad\Longleftrightarrow\quad\!\boldsymbol{\phi}^t_c=\boldsymbol{\mu}_c^t+\text{Chol}({\boldsymbol{\Sigma}}_{c}^{t})\mathbf{v}\nonumber\\
& \qquad\qquad\qquad\qquad\qquad\qquad\!\!\!\!\text{where}\quad \mathbf{v}\sim\mathcal{N}(\mathbf{0}, \mIdent),
\label{eq:pseudo}
\end{align}
where $\text{Chol}(\cdot)$ is the Cholesky decomposition\footnote{
Cholesky decomposition has a fast implementation in PyTorch and stable numerical results compared to the SVD-based matrix square root.} 
of the symmetric positive definite matrix.  

\subsection{Integration of Class Learning and the TRIPS Loss}
\label{sec:triplet_revisited}
To further preserve old class information and balance the extreme data imbalance (exemplar-free approach), during the new sessions, we want to ensure the classifier and TRIPS model have access to pseudo-samples obtained by Eq. \eqref{eq:pseudo} in order to update classifier parameters, keep features of different classes separated (not only new class features but also old class features despite the lack of access to old class samples) while actively suppressing the domain-wise information. 
Thus, the classification loss in Eq. \eqref{eq: cross_entropy_loss} is merged at the final stage with the classification loss on pseudo-features in Eq.  \eqref{eq: updated_cross_entropy_loss}.

Let $(\boldsymbol{\phi},y)\in\mathcal{S}$ be pseudo-samples (with labels $y\in\mathcal{C}''$) obtained by Eq. \eqref{eq:pseudo} for session $t$ and batch $\mathcal{B}$. We define: 
\begin{equation}
    \!\!L_{t}^\text{pseudo} = - \frac{1}{\lvert\mathcal{S}\lvert} \sum\limits_{(\boldsymbol{\phi},y) \in \mathcal{S}} \!\!\log\!\Bigg(\frac{\exp\big(\boldsymbol{\theta}_{ty}^\top \boldsymbol{\phi})\big)}{\sum\limits_{c\in \mathcal{C}'} \exp\big(\boldsymbol{\theta}_{tc}^\top \boldsymbol{\phi})\big)}
    \Bigg).
	\label{eq: updated_cross_entropy_loss}
\end{equation}
Subsequently, the TRIPS in Eq. \eqref{eq: triplet_loss} is redefined as Eq. \eqref{eq: updated_triplet_loss}:
\begin{align}
& \!\!\!L_{t}^\text{trips\_incr} = \frac{1}{\lvert\mathcal{B}\rvert} \sum\limits_{\mathbf{x}\in B}
\bigg[\max_{\mathbf{x}'\in\mathcal{B}\setminus\{\mathbf{x}\}}\gamma^+_{\mathbf{x},\mathbf{x}'}\cdot d(\mathbf{x},\mathbf{x}')
	\label{eq: updated_triplet_loss}\\
    &\!\Big.-\! \min\Big( \min_{\mathbf{x}''\in\mathcal{B}\setminus\{\mathbf{x}\}}\gamma^-_{\mathbf{x},\mathbf{x}''}\cdot d(\mathbf{x},\mathbf{x}''),
    \min_{\boldsymbol{\phi}\in\mathcal{S}} \lVert f_t(\mathbf{x})-\boldsymbol{\phi}\rVert_2^2
    \Big) + m\;\bigg]_+\!\!.\nonumber
\end{align}
Notice that negative-pair labels $\gamma^-$ between $\mathbf{x}\in\mathcal{B}$ and pseudo-samples $\boldsymbol{\phi}\in\mathcal{S}$ always equal $1$. 
Pseudo-samples are domain agnostic by design, and pseudo-sample set $\mathcal{S}$ and batch set $\mathcal{B}$ have class set $\mathcal{C}''\cap \mathcal{C}_t=\varnothing$. 
Also, to balance the old and new class performance, we sample the same number of pseudo-features as new class features $\lvert\mathcal{S}\lvert = \lvert\mathcal{B}\rvert$.

\vspace{0.1cm}
\noindent\textbf{The full loss.} $L_t$ for step $t\geq 0$ and hyper-parameters $\lambda$ \& $\lambda'$ is:
\begin{equation}
	\!\!L_t \!=\!
	\begin{cases}
	L_{0}^\text{class} \!+\! \lambda L_0^\text{trips\_base} ,&\mbox{if }  t= 0,\\
	L_t^\text{class} \!+\! L_t^\text{pseudo} \!+\! \lambda L_t^\text{trips\_incr}\! +\! \lambda' L_t^\text{dist}&\mbox{if } t\geq 1,
	\end{cases}
	\label{eq: total_loss}
\end{equation}
where $\lambda$ and $\lambda'$ are hyper-parameters that balance different loss terms. 
By diagnosing the loss value for each loss term, we set $\lambda=1$ and $\lambda'$ to 30 for all our experiments to ensure that all loss terms contribute similar level of penalty.
\begin{figure}[t]
  \hspace{-0.25cm}
   \includegraphics[width=1.05\linewidth]{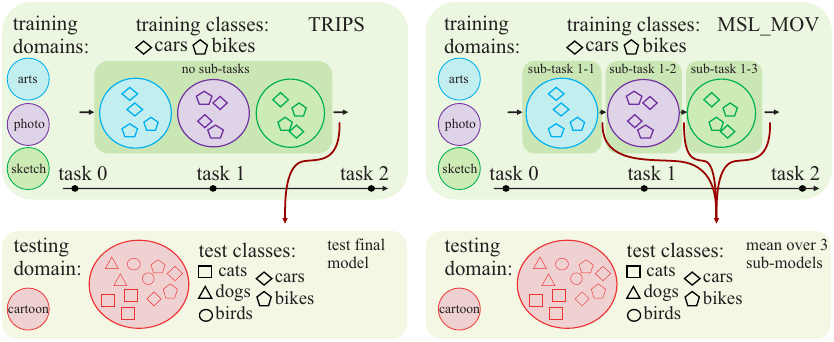}
   \vspace{-0.6cm}
    \caption{Evaluation protocols for DGCIL. ({\em right}) The only existing protocol, implemented by  MSL\_MOV, evaluates each incremental step by averaging performance on individual sub-tasks. As the first sub-task may be biased towards old classes from the previous incremental step, we argue that a better evaluation protocol should avoid sub-tasks. Also, the
first sub-task  may be more aligned with the test domain than the whole task, so evaluation should take place at the end of task instead. ({\em left}) TRIPS performs incremental  training step without splitting the problem into sub-tasks. We evaluate the samples from the test domain once the incremental step finishes. We adopt this evaluation protocol on TRIPS and MSL\_MOV.}
	\label{fig1: DGCIL_eval_prot}
\end{figure}

\section{Experiments}
\label{sec: Experiment}

In this section, we compare our proposed method with existing approaches in both incremental learning and domain generalization. We start by introducing the experimental details.

\vspace{0.1cm}
\noindent\textbf{Datasets.} 
Three domain generalization benchmark datasets PACS \citep{li2017deeper}, OfficeHome \citep{venkateswara2017deep}, and DomainNet \citep{saito2019semi} are used for our experiments.
PACS \citep{li2017deeper} contains 7 classes from 4 domains: Art, Cartoon, Photo, and Sketch.
It provides challenging recognition scenarios with significant shifting between different domains. 
OfficeHome \citep{venkateswara2017deep} is a large-scale dataset that contains 65 classes from 4 domains: Real, Clipart, Art, and Product.
DomainNet \citep{saito2019semi} is also a large-scale dataset that contains 126 classes from 4 domains: Real, Clipart, Painting, and Sketch. 
To comprehensively evaluate the performance of our proposed method, we perform various multi-step incremental learning scenarios on these datasets.
Specifically, according to the number of classes of the dataset, we perform a 2-step incremental learning task on PACS and both 5-step and 10-step incremental tasks on OfficeHome and DomainNet, similar to \cite{simon2022generalizing}.

\vspace{0.1cm}
\noindent\textbf{Evaluation Metric.} 
Domain-wise average accuracy is normally used for domain generalization tasks to evaluate the generalization capability of the model \citep{gulrajani2020search}.
Under DGCIL conditions, we also need to evaluate the memorizing capability of the model and consider performance on old and new classes, respectively. 
Thus, during each learning session, we first regard one domain as the unseen domain and the remaining domains as training domains to train the model.
Then the domain-wise class-wise average accuracy is reported as the overall performance of the model.
We also use the domain-wise harmonic accuracy between old and new classes to evaluate for possible prediction bias. 
Each domain becomes a test domain at once, \eg, given four domains we have four separate sets of experiments to average over. Moreover, our evaluation protocol is explained in Figure \ref{fig1: DGCIL_eval_prot} (left). We argue that the protocol of MSL\_MOV (right) may be biased to overestimate the performance on old classes. This may occur due to accuracy being averaged over sub-tasks, or it might even result in an overestimation of overall accuracy if early sub-tasks align more closely with the test domain than the whole task. To address this concern, we propose evaluating all models at the end of each incremental session.


\begin{figure}[t]
    \vspace{-0.3cm}
	\centering
    \begin{minipage}[t]{0.4925\linewidth}
	\includegraphics[trim={9.5cm 2cm 7cm 2cm},clip,width=1.025\linewidth]{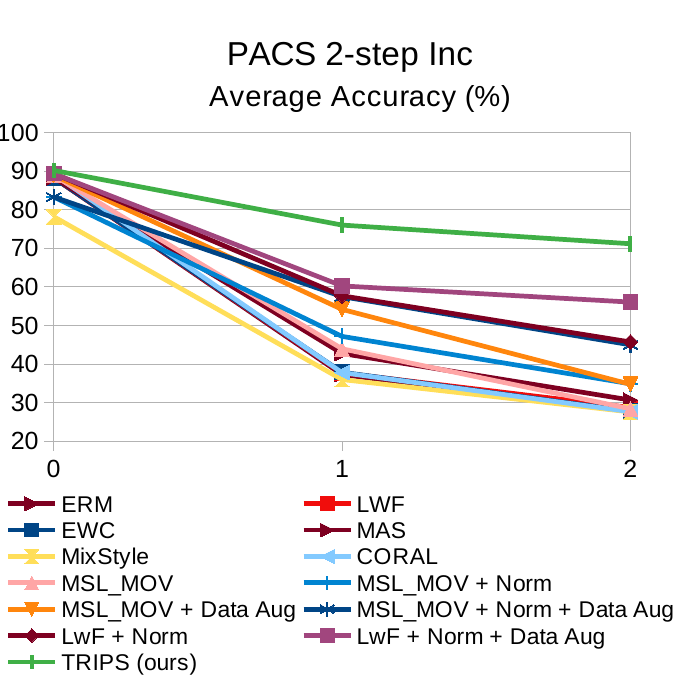}
	\end{minipage}
    \begin{minipage}[t]{0.4925\linewidth}
    \includegraphics[trim={9.5cm 2cm 7cm 2cm},clip,width=1.025\linewidth]{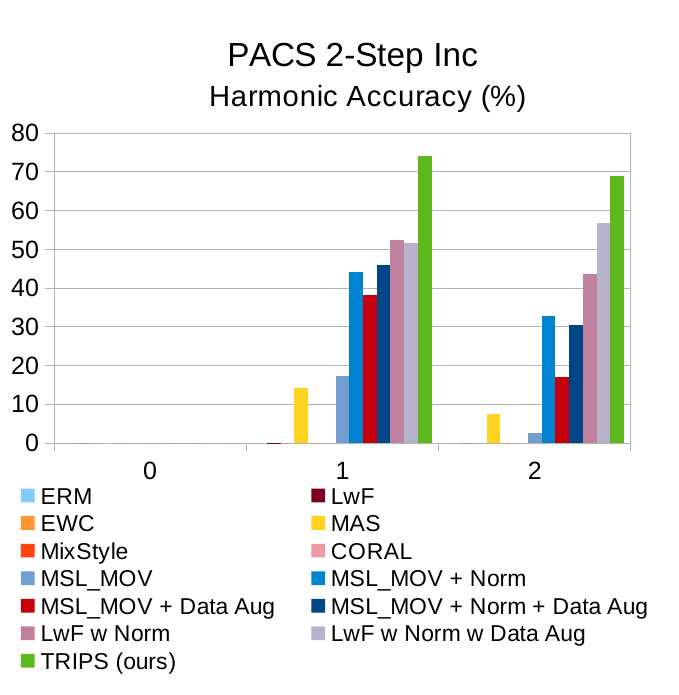}
    \label{fig: PACS_2_step_harm_acc}
    \end{minipage}
    \vspace{-0.7cm}
    \caption{Domain-average performance of each incremental step under {\em the 2-step incremental setting} on the PACS dataset.}
    \label{fig: PACS_2_step}
\end{figure}

\vspace{0.1cm}
\noindent\textbf{Model Selection.}
As the model selection is non-trivial for domain generalization, following domain generalization benchmark DomainBed \citep{gulrajani2020search}, we use the training-domain validation set strategy to select the final model for all the experiments.
Specifically, we split each training domain into 80\% and 20\% for training and validation subsets, respectively. 
We combine the validation subsets from all training domains to generate the overall validation set.
The model that provides the maximal domain-wise class-wise accuracy on the overall validation set is chosen as the final model.

\vspace{0.1cm}
\noindent\textbf{Implementation Details.} 
We build our code on top of domain generalization benchmark DomainBed \citep{gulrajani2020search}. 
For our experiments, we use ResNet-34 \citep{he2016deep} as the backbone network.  
The initial learning rate is set to 5e-5 for all dataset training. 
The maximum iteration is set to 5000 and we utilize the training-domain validation set strategy \citep{gulrajani2020search} to select the final model.
We use a batch size of 32 for each training domain data and the final data for each batch training is the cascaded data from all training domains. 
To capture the possible performance variations due to data split and report accurate performance, we run each experiment on three seeds and report the average result.

\begin{table}[!t]
\centering
\setlength\tabcolsep{3pt}
\footnotesize{
	\caption{Domain-wise average performance for each unseen domain over all incremental steps on the PACS dataset under \emph{the 2-step incremental setting.}
    }
    \vspace{-0.2cm}
	\begin{tabular}{l c c c c}
		\hline
		\multirow{3}*{Methods} & \multicolumn{4}{c}{PACS} \\
        & \multicolumn{4}{c}{2-Step Inc (+ 2 cls/step)} \\
        \cline{2-5}
        & Art & Cartoon & Photo & Sketch \\
        \hline
        & \multicolumn{4}{c}{average accuracy (\% $\uparrow$)} \\
        \hline
        ERM & 51.01 & 50.22 & 54.35 & 48.85 \\
        MixStyle \citep{zhou2021domain} & 44.68 & 47.05 & 49.15 & 47.06 \\
        CORAL \citep{sun2016deep} & 50.32 & 51.26 & 54.50 & 50.20 \\
        EWC \citep{kirkpatrick2017overcoming} & 52.01 & 50.03 & 54.57 & 50.25 \\
        MAS \citep{aljundi2018memory} & 54.52 & 53.00 & 58.48 & 51.35 \\
        MSL\_MOV \citep{simon2022generalizing} & 56.76 & 52.55 & 56.79 & 48.96 \\
        MSL\_MOV + Norm & 54.40 & 53.45 & 60.76 & 51.92 \\
        MSL\_MOV + Data Aug & 57.56 & 55.99 & 70.22 & 53.99 \\
        MSL\_MOV + Norm + Data Aug & 64.11 & 49.59 & 70.84 & \underline{63.20} \\
        MSL\_MOV + Old Prototype & 57.18 & 53.24 & 57.81 & 49.01 \\
        LwF \citep{li2017learning} & 51.71 & 50.11 & 54.43 & 47.52 \\
        LwF + Norm \citep{li2017learning} & 63.91 & 64.79 & 68.67 & 59.80 \\
        LwF + Norm + Data Aug & \underline{69.43} & \underline{72.62} & \underline{76.63} & 55.60 \\
        \rowcolor{LightCyan} TRIPS (ours) + Data Aug & \textbf{80.79} & \textbf{74.80} & \textbf{83.77} & \textbf{77.27} \\
        \hline
        & \multicolumn{4}{c}{harmonic accuracy (\% $\uparrow$)} \\
        \hline
        ERM & 0.0 & 0.0 & 0.0 & 0.0 \\
        MixStyle \citep{zhou2021domain} & 0.0 & 0.0 & 0.0 & 0.0 \\
        CORAL \citep{sun2016deep} & 0.0 & 0.0 & 0.0 & 0.0 \\
        EWC \citep{kirkpatrick2017overcoming} & 0.51 & 0.12 & 1.10 & 0.0 \\
        MAS \citep{aljundi2018memory} & 11.71 & 10.81 & 15.84 & 5.13 \\
        MSL\_MOV \citep{simon2022generalizing} & 19.90 & 8.76 & 10.29 & 0.55 \\
        MSL\_MOV + Norm & 32.71 & 40.28 & 42.11 & 38.61 \\
        MSL\_MOV + Data Aug & 22.11 & 26.90 & 49.21 & 12.39 \\
        MSL\_MOV + Norm + Data Aug & 41.19 & 40.30 & 30.19 & 40.96 \\
        MSL\_MOV + Old Prototype & 19.82 & 9.12 & 11.07 & 1.01 \\
        LwF \citep{li2017learning} & 0.0 & 0.11 & 0.05 & 0.0 \\
        LwF + Norm \citep{li2017learning} & 47.33 & 50.60 & 52.06 & \underline{42.11} \\
        LwF + Norm + Data Aug & \underline{59.02} & \underline{66.85} & \underline{66.52} & 24.35 \\
        \rowcolor{LightCyan} TRIPS (ours) + Data Aug & \textbf{74.16} & \textbf{68.04} & \textbf{73.58} & \textbf{70.47} \\
        \hline
	\end{tabular}	
	\label{tab: PACS_2_step}
}
\end{table}

\begin{table*}[!t]
\vspace{-0.3cm}
\centering
\setlength\tabcolsep{1.75pt}
\footnotesize{
	\caption{Domain-wise average performance over all incremental steps on  OfficeHome and DomainNet under \emph{the 5-step and 10-step incremental settings.}}
    \vspace{-0.2cm}
	\begin{tabular}{l c c c c c c c c c c c c c c c c c c c c}
		\hline
		\multirow{3}*{Methods} & \: & \multicolumn{8}{c}{OfficeHome} & \multicolumn{8}{c}{DomainNet} \\
        \cline{2-21}
        & \: & \multicolumn{4}{c}{5-Step Inc (+ 10 cls/step)} & \: & \multicolumn{4}{c}{10-Step Inc (+ 5 cls/step)} & \: & \multicolumn{4}{c}{5-Step Inc (+ 20 cls/step)} & \: & \multicolumn{4}{c}{10-Step Inc (+ 10 cls/step)} \\
        & \: & Art & Clipart & Product & Real & \: & Art & Clipart & Product & Real & \: & Clipart & Painting & Real & Sketch & \: & Clipart & Painting & Real & Sketch \\
        \hline
        & \: & \multicolumn{19}{c}{average accuracy (\% $\uparrow$)} \\
        \hline
        ERM & \: & 26.55 & 26.85 & 31.95 & 32.41 & \: & 16.19 & 16.19 & 18.81 & 19.49 & \: & 30.60 & 27.69 & 30.64 & 30.09 & \: & 18.11 & 16.59 & 18.04 & 17.79 \\
        MixStyle \citep{zhou2021domain}$\!\!\!\!$ & \: & 15.16 & 18.96 & 22.33 & 21.87 & \: & 10.15 & 12.27 & 14.04 & 14.32 & \: & 21.12 & 16.85 & 17.98 & 20.64 & \: & 13.21 & 10.44 & 11.79 & 13.27 \\
        CORAL \citep{sun2016deep}$\!\!\!\!$ & \: & 27.39 & 26.91 & 31.52 & 32.34 & \: & 16.12 & 15.79 & 18.31 & 19.11 & \: & 30.32 & 28.45 & 31.05 & 30.06 & \: & 18.12 & 16.77 & 18.15 & 17.87 \\
        EWC \citep{kirkpatrick2017overcoming}$\!\!\!\!$ & \: & 26.85 & 27.00 & 32.60 & 31.71 & \: & 16.07 & 16.94 & 18.53 & 19.34 & \: & 30.45 & 27.95 & 30.45 & 30.01 & \: & 18.11 & 16.57 & 18.24 & 17.94 \\
        MAS \citep{aljundi2018memory}$\!\!\!\!$ & \: & 29.95 & 30.56 & 41.83 & 36.51 & \: & 19.32 & 21.28 & 22.51 & 24.01 & \: & 29.86 & 28.26 & 31.86 & 30.22 & \: & 18.14 & 16.37 & 19.74 & 17.99 \\
        MSL\_MOV \citep{simon2022generalizing}$\!\!\!\!$ & \: & 30.43 & 30.65 & 36.62 & 39.29 & \: & 19.68 & 20.81 & 23.81 & 27.05 & \: & 33.52 & 31.69 & 37.87 & 33.64 & \: & 20.66 & 19.59 & 23.00 & 21.08 \\
        LwF \citep{li2017learning}$\!\!\!\!$ & \: & 26.82 & 26.95 & 32.15 & 33.02 & \: & 16.07 & 16.81 & 19.21 & 20.59 & \: & 30.60 & 28.05 & 29.89 & 30.08 & \: & 17.91 & 15.88 & 18.64 & 17.82 \\
        LwF + Norm & \: & 38.06 & \underline{41.17} & \underline{47.77} & \underline{51.46} & \: & 25.83 & \underline{28.37} & \underline{34.45} & \underline{39.21} & \: & \underline{36.25} & \underline{34.62} & \underline{40.41} & \underline{39.32} & \: & 29.10 & \underline{26.81} & 30.84 & 29.06 \\
        LwF+Norm+Data Aug & \: & \underline{40.03} & 37.92 & 46.98 & 49.02 & \: & \underline{30.85} & 27.66 & 33.59 & 34.98 & \: & 35.92 & 31.59 & 35.20 & 35.60 & \: & \underline{31.62} & 26.02 & \underline{31.10} & \underline{29.78} \\
        \rowcolor{LightCyan} TRIPS (ours) & \: & \textbf{46.04} & \textbf{48.99} & \textbf{62.24} & \textbf{67.88} & \: & \textbf{33.87} & \textbf{38.64} & \textbf{49.57} & \textbf{54.39} & \: & \textbf{54.20} & \textbf{47.36} & \textbf{55.35} & \textbf{55.22} & \: & \textbf{47.80} & \textbf{42.41} & \textbf{50.37} & \textbf{49.41} \\
        \hline
        & \: & \multicolumn{19}{c}{harmonic accuracy (\% $\uparrow$)} \\
        \hline
        ERM & \: & 0.63 & 0.84 & 0.14 & 0.0 & \: & 0.10 & 0.33 & 0.48 & 1.25 & \: & 0.15 & 0.08 & 0.03 & 0.08 & \: & 0.09 & 0.10 & 0.01 & 0.04 \\
        MixStyle \citep{zhou2021domain}$\!\!\!\!$ & \: & 0.0 & 0.01 & 0.0 & 0.0 & \: & 0.0 & 0.0 & 0.04 & 0.0 & \: & 0.01 & 0.0 & 0.02 & 0.0 & \: & 0.0 & 0.0 & 0.02 & 0.0 \\
        CORAL \citep{sun2016deep}$\!\!\!\!$ & \: & 0.0 & 0.0 & 0.02 & 0.01 & \: & 0.0 & 0.0 & 0.1 & 0.2 & \: & 0.0 & 0.0 & 0.01 & 0.0 & \: & 0.0 & 0.05 & 0.01 & 0.01 \\
        EWC \citep{kirkpatrick2017overcoming}$\!\!\!\!$ & \: & 0.24 & 0.13 & 2.32 & 0.03 & \: & 0.09 & 0.54 & 0.0 & 0.02 & \: & 0.20 & 0.02 & 0.30 & 0.0 & \: & 0.06 & 0.11 & 0.12 & 0.11 \\
        MAS \citep{aljundi2018memory} & \: & 5.60 & 16.85 & 20.08 & 10.65 & \: & 7.03 & 12.17 & 7.68 & 7.26 & \: & 4.59 & 4.20 & 5.15 & 2.89 & \: & 3.32 & 2.15 & 4.69 & 3.10 \\
        MSL\_MOV \citep{simon2022generalizing}$\!\!\!\!$ & \: & 10.69 & 11.41 & 14.75 & 18.23 & \: & 7.47 & 9.70 & 11.40 & 15.28 & \: & 13.31 & 12.58 & 18.20 & 12.68 & \: & 8.42 & 8.13 & 11.20 & 8.85 \\
        LwF \citep{li2017learning}$\!\!\!\!$ & \: & 0.23 & 0.46 & 0.0 & 0.0 & \: & 0.10 & 0.08 & 0.16 & 0.25 & \: & 0.15 & 0.10 & 0.25 & 0.12 & \: & 0.09 & 0.12 & 0.03 & 0.01 \\
        LwF + Norm & \: & 31.14 & \underline{37.02} & \underline{40.49} & \underline{44.36} & \: & 20.84 & \underline{26.88} & \underline{32.35} & \underline{37.04} & \: & \underline{22.71} & \underline{22.30} & \underline{28.30} & \underline{25.58} & \: & 26.62 & \underline{24.20} & 29.36 & 25.83 \\
        LwF+Norm+Data Aug & \: & \underline{34.00} & 31.87 & 39.05 & 41.63 & \: & \underline{29.72} & 26.19 & 31.77 & 33.75 & \: & 22.68 & 19.71 & 22.88 & 22.77 & \: & \underline{30.05} & \underline{24.20} & \underline{29.98} & \underline{28.19}  \\
        \rowcolor{LightCyan} TRIPS (ours) & \: & \textbf{39.56} & \textbf{44.89} & \textbf{57.46} & \textbf{63.71} & \: & \textbf{32.40} & \textbf{39.07} & \textbf{49.68} & \textbf{55.10} & \: & \textbf{42.88} & \textbf{36.97} & \textbf{44.50} & \textbf{43.28} & \: & \textbf{49.93} & \textbf{44.49} & \textbf{52.45} & \textbf{50.59} \\
        \hline
	\end{tabular}	
	\label{tab: OfficeHome_DomainNet}
}
\end{table*}

\begin{figure*}[!t]
	\centering
    \begin{minipage}[c]{0.2225\linewidth}
	\includegraphics[trim={9cm 3cm 7.5cm 3cm},clip,width=1.05\linewidth]{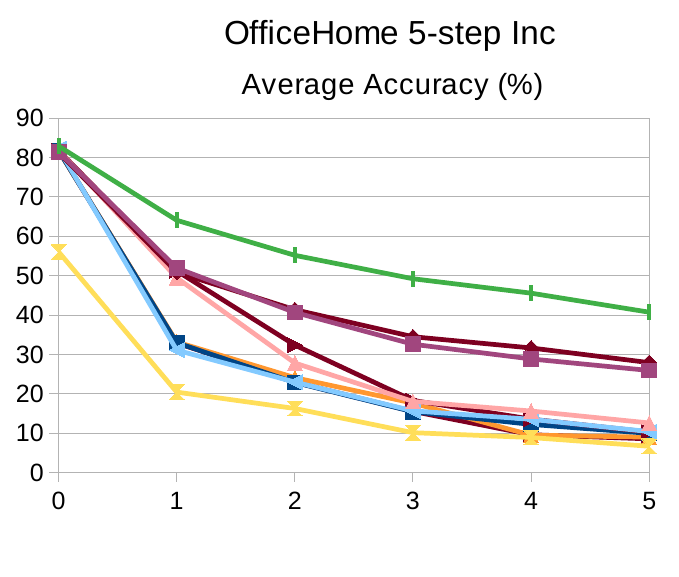}
    \label{fig: OfficeHome_5_step_acc}
	\end{minipage}
    \begin{minipage}[c]{0.2225\linewidth}
    \includegraphics[trim={9cm 3cm 7.5cm 3cm},clip,width=1.05\linewidth]{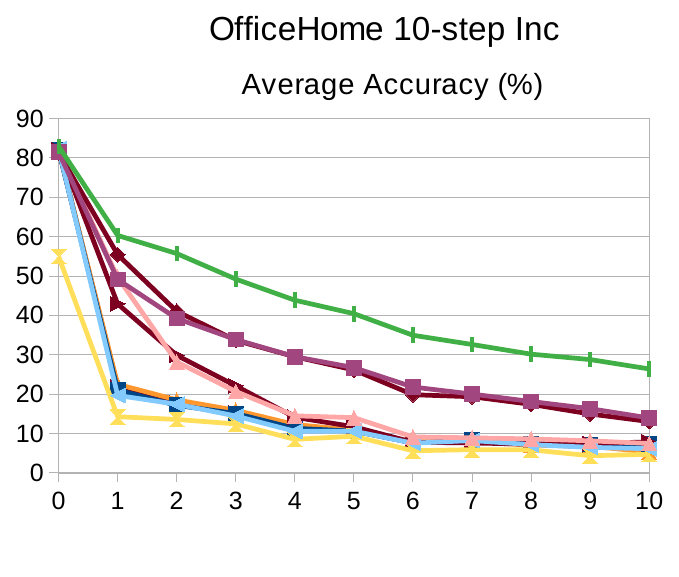}
    \label{fig: OfficeHome_10_step_acc}
    \end{minipage}
    \begin{minipage}[c]{0.2225\linewidth}
	\includegraphics[trim={9cm 3cm 7.5cm 3cm},clip,width=1.05\linewidth]{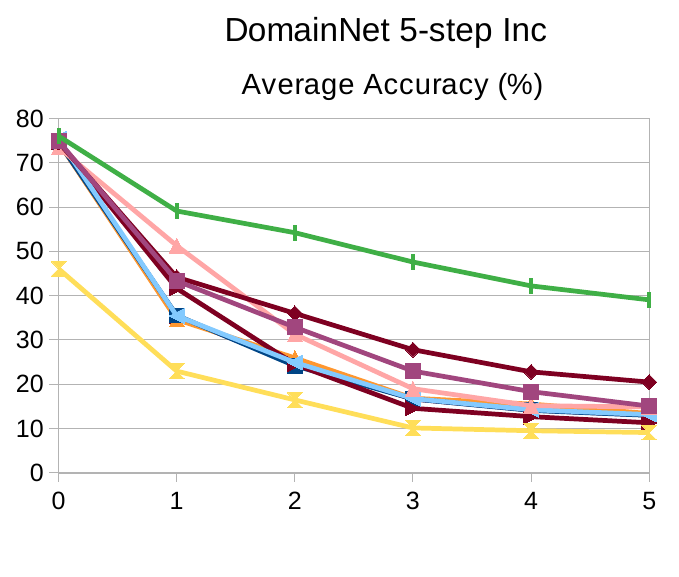}
    \label{DomainNet_5_step_acc}
	\end{minipage}
    \begin{minipage}[c]{0.2225\linewidth}
    \includegraphics[trim={9cm 3cm 7.5cm 3cm},clip,width=1.05\linewidth]{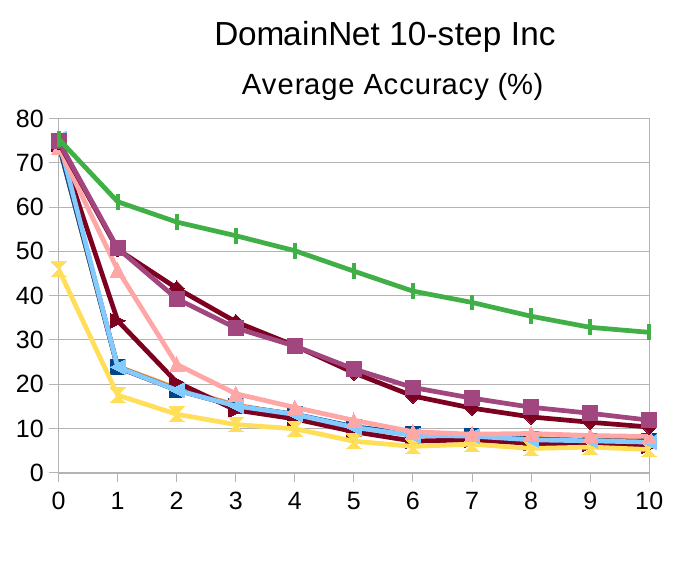}
    \label{fig: DomainNet_10_step_acc}
    \end{minipage}
    \begin{minipage}[c]{0.09\linewidth}
    \includegraphics[trim={17.625cm 5cm 6.5cm 5cm},clip,width=1.125\linewidth]{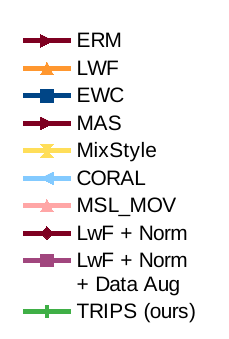}
    \label{fig: acc_label}
    \end{minipage}
    \begin{minipage}[c]{0.2225\linewidth}
    \vspace{-1cm}
	\includegraphics[trim={9cm 3cm 7.5cm 3cm},clip,width=1.05\linewidth]{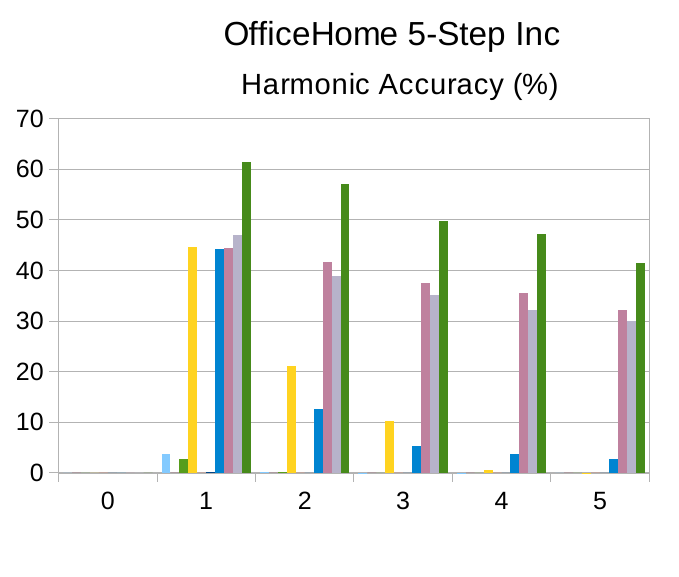}
    \label{fig: OfficeHome_5_step_harm_acc}
	\end{minipage}
    \begin{minipage}[c]{0.2225\linewidth}
    \vspace{-1cm}
    \includegraphics[trim={9cm 3cm 7.5cm 3cm},clip,width=1.05\linewidth]{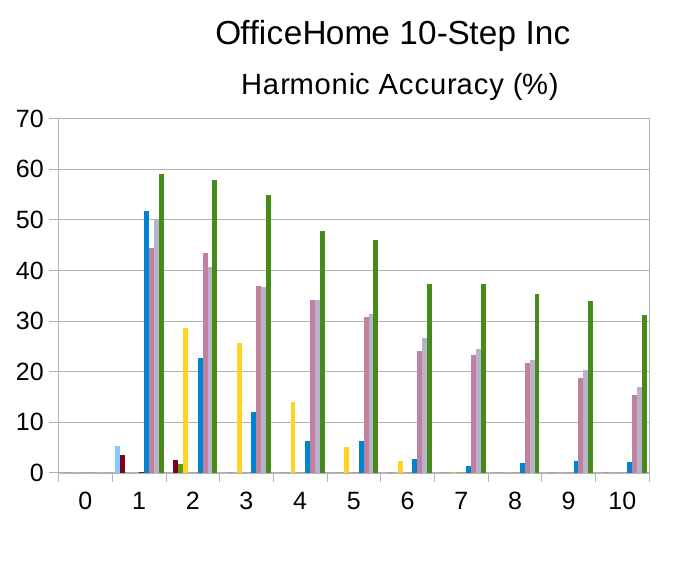}
    \label{fig: OfficeHome_10_step_harm_acc}
    \end{minipage}
    \begin{minipage}[c]{0.2225\linewidth}
    \vspace{-1cm}
	\includegraphics[trim={9cm 3cm 7.5cm 3cm},clip,width=1.05\linewidth]{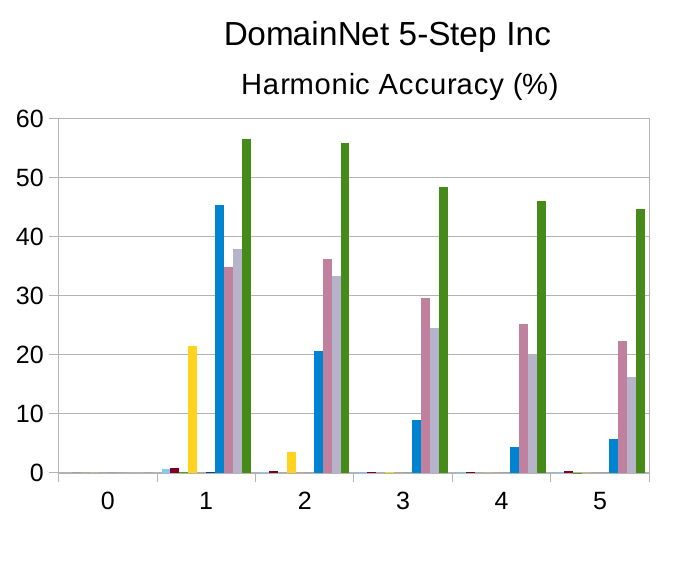}
    \label{fig: DomainNet_5_step_harm_acc}
	\end{minipage}
    \begin{minipage}[c]{0.2225\linewidth}
    \vspace{-1cm}
    \includegraphics[trim={9cm 3cm 7.5cm 3cm},clip,width=1.05\linewidth]{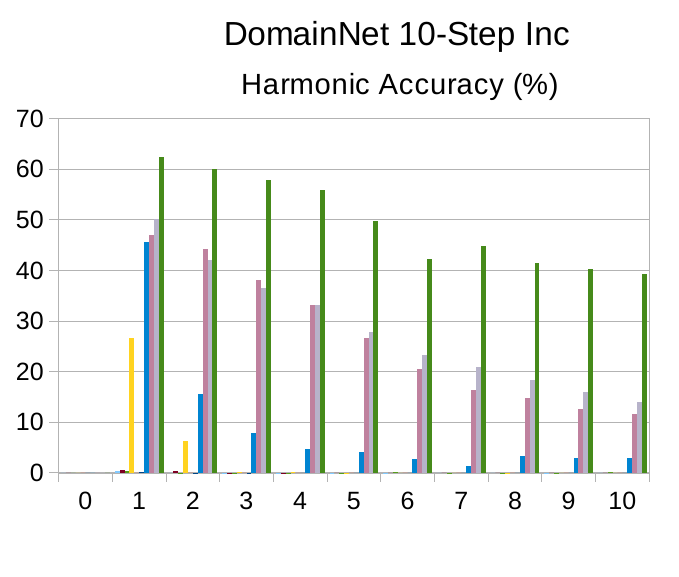}
    \label{fig: DomainNet_10_step_harm_acc}
    \end{minipage}
    \begin{minipage}[c]{0.09\linewidth}
    \includegraphics[trim={16.5cm 2.5cm 6.5cm 5.5cm},clip,width=1.5\linewidth]{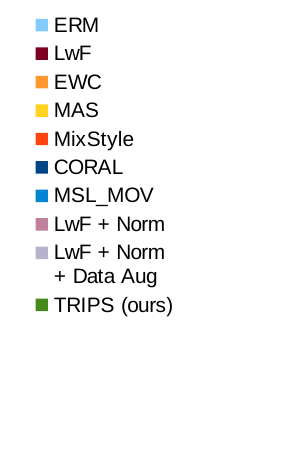}
    \label{fig: harm_acc_label}
    \end{minipage}
    \setlength{\abovecaptionskip}{-1.3cm}
    \caption{Domain-average performance of each incremental step under {\em the 5-step and 10-step incremental settings} on the OfficeHome and DomainNet datasets.}
    \label{fig: OfficeHome_DomainNet}
\end{figure*}

\subsection{Comparison with the State of the Art}
We compare our method with exemplar-free incremental learning methods LwF \citep{li2017learning}, EWC \citep{kirkpatrick2017overcoming}, and MAS \citep{aljundi2018memory}.
We also compare our method with domain generalization methods CORAL \citep{sun2016deep} and MixStyle \citep{zhou2021domain}, and DGCIL method MSL\_MOV \citep{simon2022generalizing}.
Experiments are performed on three datasets \citep{li2017deeper, venkateswara2017deep, saito2019semi} under several different incremental settings.
DGCIL is a challenging task that combines the difficulty of both incremental learning and domain generalization. 
To comprehensively evaluate the performance of each method, its memorizing and generalization capabilities should be evaluated. 
Thus, we report the average performance towards all incremental steps for each domain to evaluate the out-of-distribution performance. 
We also report the average performance towards all domains for each incremental step to evaluate the preservation of past knowledge.  

Figure \ref{fig: PACS_2_step} and Table \ref{tab: PACS_2_step} show the experimental results on the PACS dataset. 
During experiments, we find that applying batch normalization on the features from the feature extractor is important to boost performance.
We conjecture this is because, during training, there are covariate shifts \citep{ioffe2015batch}, and the shifting is more severe when the training data comes from various domains.
To make fair comparisons, we apply the same batch normalization to all distillation-based methods. 
In addition, for small dataset training, data augmentation is important to guarantee sufficient data is available for training. 
Thus, we apply image rotations by 90 degrees on the training data and regard the rotated images as additional classes to increase both the number of samples and class concepts.  
In each incremental step,  we apply this data and class augmentation strategy in our method when the number of new classes is below 5. 
For fair comparisons, we apply the same augmentation strategy to our method and other methods.

We also note that MSL\_MOV is an exemplar-based method by design. However, we study the exemplar-free setting in this paper. Thus, for a fair comparison, we equip MSL\_MOV  with mean-based prototypes and we completely remove exemplars from MSL\_MOV\footnote{Prototype-based IL methods usually perform worse than exemplar-based IL methods.  Thus, the performance of MSL\_MOV reported by \cite{simon2022generalizing} differs from the performance reported in our paper due to (i) prototype-based ``memory'' and a more realistic evaluation setting as in Figure \ref{fig1: DGCIL_eval_prot} (left).}.

According to Table \ref{tab: PACS_2_step}, TRIPS always achieves the best performance on different unseen test domains. 
We have an average of 8.69\% and 12.94\% gain on average and harmonic accuracy compared to the second-best method, respectively.  
When the hardest {\em sketch} domain is the unseen test domain, our method outperforms SOTA by 14.07\% and 28.36\% in average and harmonic accuracy, respectively.
This shows TRIPS copes well with such an out-of-distribution scenario.  
Figure \ref{fig: PACS_2_step} shows that TRIPS also achieves both the highest average accuracy and harmonic accuracy for all the incremental steps. 

Figure \ref{fig: OfficeHome_DomainNet} and Table \ref{tab: OfficeHome_DomainNet} show the experimental results on OfficeHome and DomainNet datasets.
Under both five-step and ten-step incremental settings, TRIPS shows significant improvements over the baselines. 
According to Table \ref{tab: OfficeHome_DomainNet}, under four different task settings, we outperform the second-best baseline by an average of 13.78\%/15.86\% on average/harmonic accuracy for all domains which shows our method copes well with the out-of-distribution test data.  
According to Figure \ref{fig: OfficeHome_DomainNet}, under four different task settings, we also outperform the second-best baseline in almost every step on both average and harmonic accuracy which shows our method is more robust to catastrophic forgetting. 

\subsection{Ablation Study}
To validate the effectiveness of each part of our method, Tables \ref{tab: Ablation_OfficeHome} and \ref{tab: sub_Ablation_OfficeHome} show the ablation study on the OfficeHome dataset \citep{venkateswara2017deep}.
Firstly,  when the triplet loss is used (which improves the generalization of the model), in the base step, the average accuracy of the model improves by 1.60\%. 
During incremental learning, the triplet loss improves the average/harmonic accuracy in each incremental step for every test domain by an average of 12.06\%/11.34\%. 
Thus, the benefit of triplet loss is clear.
Next, we update and inject old-class prototypes into both the cross-entropy and triplet loss. 
We find that directly injecting updated old-class prototypes does not help improve the performance.
We conjecture this is because of the extreme data imbalance  since we focus on the exemplar-free setting and each old class only has one mean-based prototype  preserved.
Thus, to overcome the data imbalance  and obtain balanced old- and new-class performance, sufficient pseudo-features of old classes are sampled from the updated class prototype representations.  
With the help of  sampling, the performance is boosted by an average  of 1.57\%/ 1.47\% average/harmonic accuracy over all incremental steps.

\begin{table*}[!t]
\setlength\tabcolsep{3pt}
\centering
\footnotesize{
	\caption{Ablations on OfficeHome under {\em the five-step incremental setting} (ten new classes added at a time). Results use format $a\: (b, c, d, e)$ where $a$ is (top) the average and (bottom) harmonic accuracy over all four test domains, whereas $(b, c, d, e)$ represents (top) the average and (bottom) harmonic accuracy per test domain: art, clipart, product, real. The detailed performance for steps 3, 4, and 5 are reported in the supplementary material due to space limitations.}
    \vspace{-0.2cm}
	\begin{tabular}{c c c c | c c c c c c}
		\hline
		\multirow{2}*{KD} & \multirow{2}*{triplet} & \multirow{2}*{\tabincell{c}{proto \\ shift}} & \multirow{2}*{\tabincell{c}{pseudo \\ sampling}} & 0 (base) & 1 & 2 & 3 & 4 & 5 \\
        \cline{5-10}
        & & & & \multicolumn{6}{c}{Domain-Average (Art, Clipart, Product, Real)} \\
        \hline
        & & & & \multicolumn{6}{c}{average accuracy (\% $\uparrow$)} \\
        \hline
        \checkmark &  &  &  & 81.29 (76.26 70.83 86.82 91.27) & 50.89 (44.71 47.99 54.12 56.73) & 41.37 (33.12 38.64 46.38 47.35) & 34.52 & 31.67 & 27.94 \\
        \checkmark & \checkmark &  &  & \textbf{82.89 (79.33 72.20 87.74 92.30)} & 60.85 (49.97 56.35 66.60 70.48) & 52.97 (39.22 47.42 60.48 64.74) & 47.62 & 44.81 & 40.18 \\
        \checkmark  & \checkmark & \checkmark &  & \textbf{82.89 (79.33 72.20 87.74 92.30)} & 60.73 (49.39 53.84 69.50 70.19) & 53.38 (40.61 46.17 61.50 65.26) & 48.27 & 44.68 & 39.92 \\
        \rowcolor{LightCyan} \checkmark  & \checkmark & \checkmark & \checkmark & \textbf{82.89 (79.33 72.20 87.74 92.30)} & \textbf{64.08 (50.21 58.33 71.45 76.31)} & \textbf{55.18 (42.83 48.25 62.23 67.42)} & \textbf{49.25} & \textbf{45.58} & \textbf{40.74} \\
        \hline
        & & & & \multicolumn{6}{c}{harmonic accuracy (\% $\uparrow$)} \\
        \hline
        \checkmark &  &  &  & - & 44.45 (40.32 45.81 44.98 46.71) & 41.58 (32.74 39.43 46.25 47.89) & 37.49 & 35.52 & 32.22 \\
        \checkmark & \checkmark &  &  & - & 58.06 (\textbf{43.55} 55.76 63.33 69.61) & 54.35 (38.03 49.71 62.42 67.24) & 48.83 & 46.42 & 40.30 \\
        \checkmark & \checkmark & \checkmark &  & - & 58.09 (42.81 52.07 68.16 69.30) & 55.51 (41.78 48.17 \textbf{64.13} 67.93) & 49.63 & 46.50 & 39.95 \\
        \rowcolor{LightCyan} \checkmark  & \checkmark & \checkmark & \checkmark & - & \textbf{61.45 (}43.17 \textbf{57.30 69.15 76.16)} & \textbf{57.11 (44.33 50.42} 63.94 \textbf{69.74)} & \textbf{49.76} & \textbf{47.22} & \textbf{41.51} \\
        \hline
	\end{tabular}	
	\label{tab: Ablation_OfficeHome}
}
\end{table*}

\begin{table}[!t]
\setlength\tabcolsep{4pt}
\footnotesize{\caption{Comparison of our sampling strategy with other pseudo-sampling methods, \ie, SDC \citep{yu2020semantic} and PASS \citep{zhu2021prototype} under {\em the five-step incremental setting} (10 new classes added at a time) on the OfficeHome dataset.}
\vspace{-0.2cm}
	\begin{tabular}{l|c c c c c c}
		\hline
		& 0 (base) & 1 & 2 & 3 & 4 & 5  \\
        \hline
        & \multicolumn{6}{c}{average accuracy (\% $\uparrow$)} \\
        \hline
        MSL\_MOV & 81.80 & 49.49 & 27.84 & 18.08 & 15.65 & 12.62 \\
        MSL\_MOV with SDC & 81.80 & 50.31 & 28.62 & 19.01 & 16.04 & 13.08 \\
        TRIPS w/o prototypes & 82.89 & 60.85 & 52.97 & 47.62 & 44.81 & 40.18 \\
        TRIPS with SDC & 82.89 & 60.73 & 53.38 & 48.27 & 44.68 & 39.92 \\
		TRIPS with PASS & 82.89 & 63.37 & 53.53 & 48.08 & 43.92 & 39.07 \\
		\rowcolor{LightCyan} TRIPS (ours) & 82.89 & \textbf{64.08} & \textbf{55.18} & \textbf{49.25} & \textbf{45.58} & \textbf{40.74} \\
		\hline
        & \multicolumn{6}{c}{harmonic accuracy (\% $\uparrow$)} \\
        \hline
        MSL\_MOV & - & 44.16 & 12.70 & 5.39 & 3.79 & 2.80 \\
        MSL\_MOV with SDC & - & 44.21 & 13.02 & 6.07 & 4.13 & 3.05 \\
        TRIPS w/o prototypes & - & 58.06 & 54.35 & 48.83 & 46.42 & 40.30 \\
        TRIPS with SDC & - & 58.09 & 55.51 & 49.63 & 46.50 & 39.95 \\
        TRIPS with PASS & - & 59.51 & 55.20 & 49.11 & 46.51 & 40.82 \\
		\rowcolor{LightCyan} TRIPS (ours) & - & \textbf{61.45} & \textbf{57.11} & \textbf{49.76} & \textbf{47.22} & \textbf{41.51} \\
		\hline
	\end{tabular}	
	\label{tab: OfficeHome_compare_w_other_ps}
}
\end{table}

\subsection{Comparison of Different Prototype Representations}
To validate the efficiency of our proposed prototype representation based on the multivariate Normal distribution (mean and covariance matrix per class) and our pseudo-sampling strategy, we compare our method with related prototype representation methods \citep{yu2020semantic, zhu2021prototype}.
SDC \citep{yu2020semantic} utilizes the single mean-based prototype per class and a non-parametric cosine classifier.
SDC updates old-class means by shifting them through the lens of new-class samples and uses updated class-wise means for classification.
As in SDC, the final classification result is determined by the maximum cosine similarity over the feature vector of an input image and the updated class-wise mean-based prototypes. SDC does not sample pseudo-features of old classes as it is impossible to perform sampling from first-order statistics such as the mean. 

PASS \citep{zhu2021prototype} uses the  mean-based prototype with a radius (hyper-parameter) to form an isotropic Normal distribution from which pseudo-features of old classes are sampled.
The formula 
%
$\boldsymbol{\phi}_c^t =\boldsymbol{\mu}_c^t +  \gamma\mathbf{v}$ describes the sampling step, 
%
where $\boldsymbol{\phi}_c^t$ are the sampled pseudo-features of old class $c$ at session $t$ and $\boldsymbol{\mu}_c^t$ is the  mean-based prototype for old class $c$  at session $t$. 
Moreover, $\mathbf{v}\sim \mathcal{N}(\mathbf{0}, \mIdent)$ is a vector sampled from the isotropic Normal distribution, whereas 
$\gamma\geq 0$ is the radius that controls the deviation of pseudo-features from the mean.

We implement the SDC and PASS methods in our setting using their official code available online.
As MSL\_MOV is an exemplar-based method which does not utilize old-class prototypes, for a fair comparison, we  implement it with SDC (updated old prototypes).

Table \ref{tab: OfficeHome_compare_w_other_ps} shows that  modeling prototype representations as full multivariate Normal distributions per class outperforms the naive use of isotropic Normal distributions with distinct means per class.

\begin{figure*}[h]
    \vspace{-0.2cm}
	\centering
    \begin{minipage}[t]{0.3\linewidth}
    \includegraphics[trim={9.5cm 2cm 7.5cm 3cm},clip,width=\linewidth]{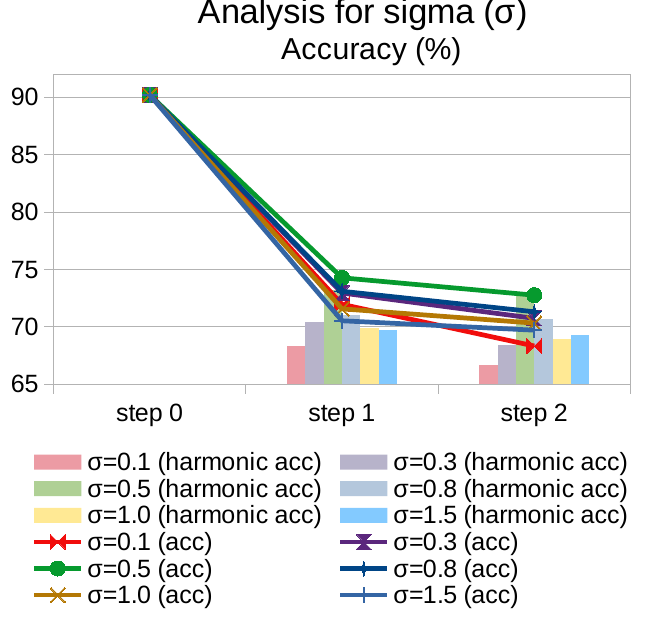}
    \label{fig: hyparam_sigma}
    \end{minipage}
    \begin{minipage}[t]{0.3\linewidth}
    \includegraphics[trim={9.5cm 2cm 7.5cm 3cm},clip,width=\linewidth]{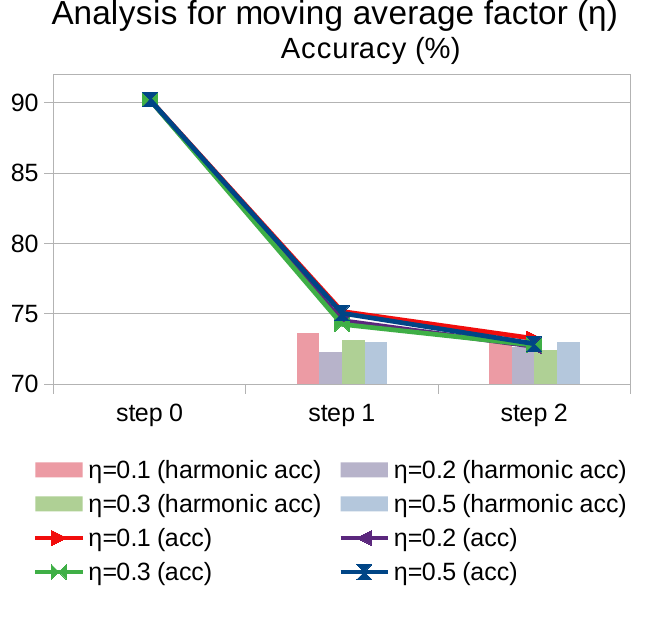}
    \label{fig: hyparam_eta}
    \end{minipage}
    \begin{minipage}[t]{0.3\linewidth}
	\includegraphics[trim={9.5cm 2cm 7.5cm 2.9cm},clip,width=\linewidth]{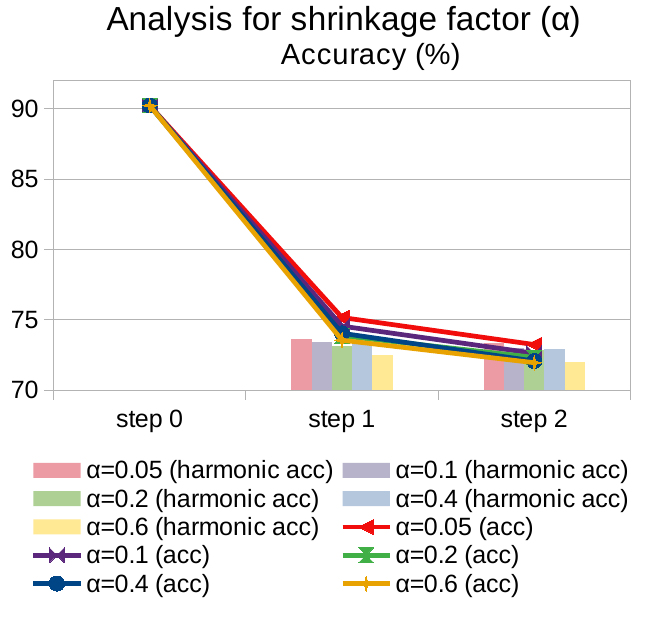}
    \label{fig: hyparam_alpha}
	\end{minipage}
    \vspace{-0.8cm}
    \caption{Ablations of prototype shifting and pseudo-sampling hyper-parameters $\sigma$, $\eta$, and $\alpha$ on the PACS dataset.}
    \label{fig: hyparam_sigma_eta_alpha}
\end{figure*}

\begin{figure}[ht]
	\centering
    \begin{minipage}[t]{0.49\linewidth}
	\includegraphics[trim={9.5cm 2cm 7.5cm 2cm},clip,width=\linewidth]{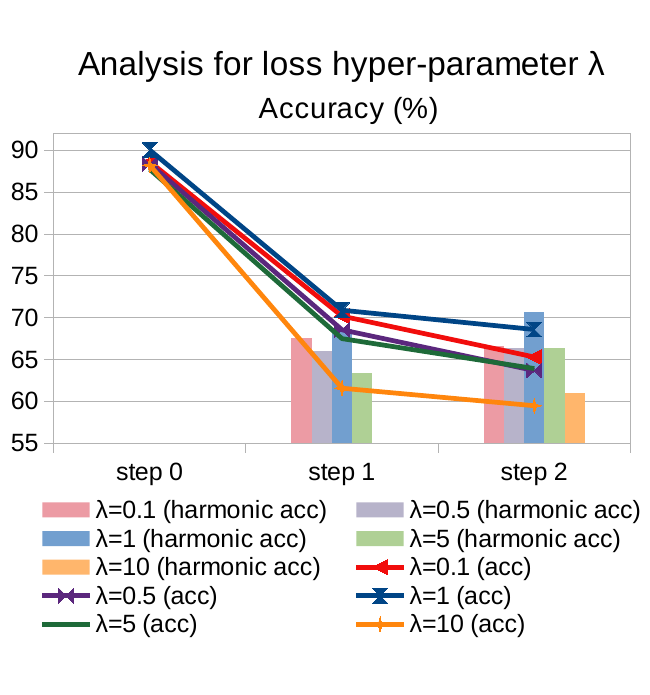}
    \label{fig: hyparam_lambda_tri}
	\end{minipage}
    \begin{minipage}[t]{0.49\linewidth}
    \includegraphics[trim={9.5cm 2cm 7.5cm 2cm},clip,width=\linewidth]{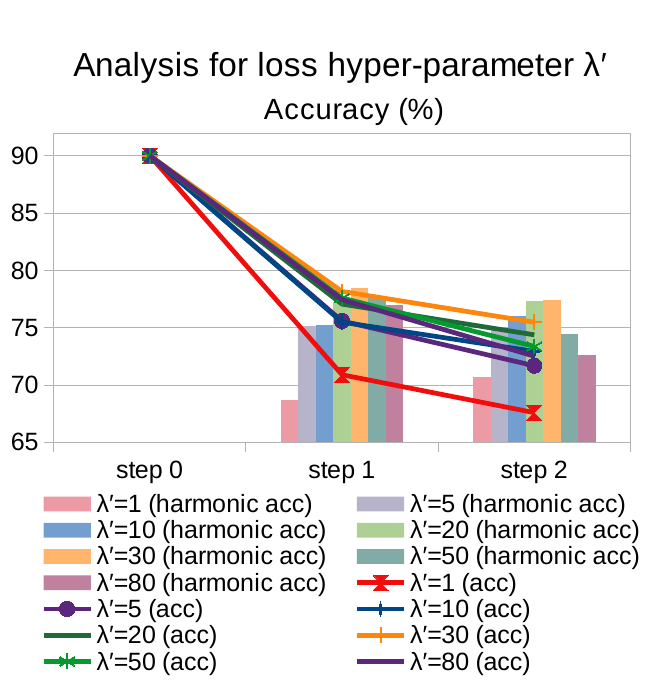}
    \label{fig: hyparam_lambda_dis}
    \end{minipage}
    \vspace{-0.8cm}
    \caption{Ablations of loss hyper-parameters $\lambda$ and $\lambda'$ on the PACS dataset.}
    \label{fig: hyparam_lambda}
\end{figure}

\subsection{Hyper-parameter Analysis}
\label{sub-sec: Hyper-parameter Analysis}
To find the most suitable hyper-parameter value, we perform the hyper-parameter grid analysis on the PACS dataset under the two-step incremental protocol.
Figure \ref{fig: hyparam_lambda} shows the analysis for different loss terms. 
To better explore the effect of each loss term, these experiments are trained without data augmentation and pseudo-sampling. 
Five old exemplars for each class and each domain are stored and utilized.
Firstly, we set the distillation loss term hyper-parameter ($\lambda'$) to 1 and vary the value for the triplet loss term hyper-parameter ($\lambda$).
We find that when $\lambda$ is set to 1, the best average and harmonic accuracy are acquired.
Then we set $\lambda$ to 1 and vary the value for $\lambda'$.
We find that when $\lambda'$ is set to 30, the best average and harmonic accuracy are acquired.
Thus, for all our experiments, we set $\lambda$ to 1 and $\lambda'$ to 30.
We do a similar loss term hyper-parameter grid search for other methods.
For all distillation-based methods \citep{li2017learning, simon2022generalizing} we use the hyper-parameter value 30 for the distillation loss.
For importance matrix-based methods \citep{kirkpatrick2017overcoming, aljundi2018memory}, we use the hyper-parameter value 1000 for the regularization loss.

We also explore the effect of various hyper-parameters related to prototype shifting and pseudo-sampling.
Figure \ref{fig: hyparam_sigma_eta_alpha} shows the 
values of hyper-parameters we set  
for shifting factor ($\sigma$), moving average factor ($\eta$), and shrinkage factor ($\alpha$).
Firstly, we set $\eta=0.1$ and $\alpha=0.1$, and vary the value of $\sigma$. 
When $\sigma$ is set to 0.5, the best performance is obtained.
Then, we set $\sigma=0.5$ and $\alpha=0.1$, and vary the value of $\eta$. 
When $\eta$ is set to 0.1, the best performance is obtained.
After that, we set $\sigma=0.5$ and $\eta=0.1$, and vary the value of $\alpha$.
When $\alpha$ is set to 0.05, the best performance is obtained.
These parameters were selected on the PACS dataset and then simply applied without any tuning on both OfficeHome and DomainNet datasets. 
We also observed that the same hyper-parameters were optimal when we used the validation set of PACS comprised of the three training domains of PACS and 20\% train data withheld for validation.

\begin{figure}[t]
    \vspace{-0.3cm}
	\centering
    \begin{minipage}[t]{0.47\linewidth}
	\includegraphics[trim={9cm 2cm 7.5cm 2cm},clip,width=\linewidth]{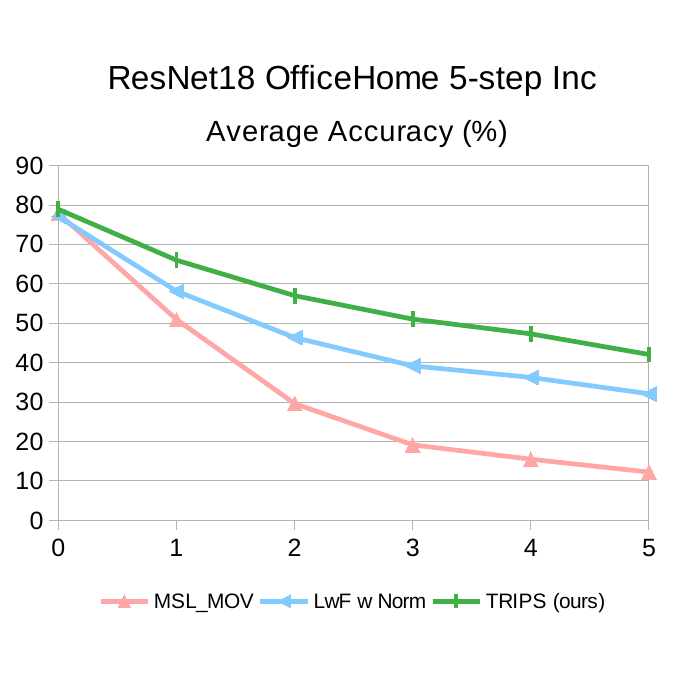}
    \label{fig: ResNet18_OfficeHome_5_step_acc}
	\end{minipage}
    \begin{minipage}[t]{0.47\linewidth}
    \includegraphics[trim={9cm 2cm 7.5cm 2cm},clip,width=\linewidth]{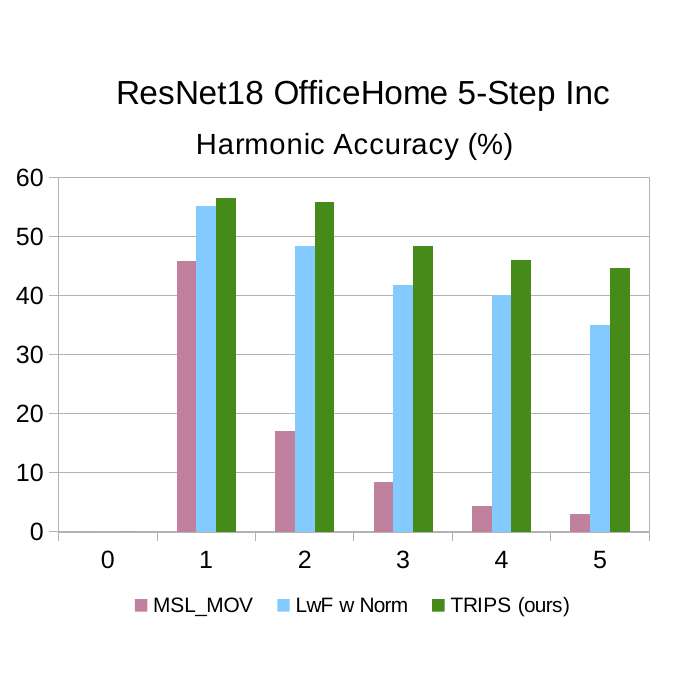}
    \label{fig: ResNet18_OfficeHome_5_step_harm_acc}
    \end{minipage}
    \vspace{-0.8cm}
    \caption{Performance under {\em the five-step incremental settings} (10 new classes added at a time) on the OfficeHome dataset based on ResNet-18.}
    \label{fig: ResNet18_OfficeHome}
\end{figure}

\begin{figure}[t]
    \vspace{-0.3cm}
	\centering
    \begin{minipage}[t]{0.47\linewidth}
	\includegraphics[trim={9cm 2cm 7.5cm 2cm},clip,width=\linewidth]{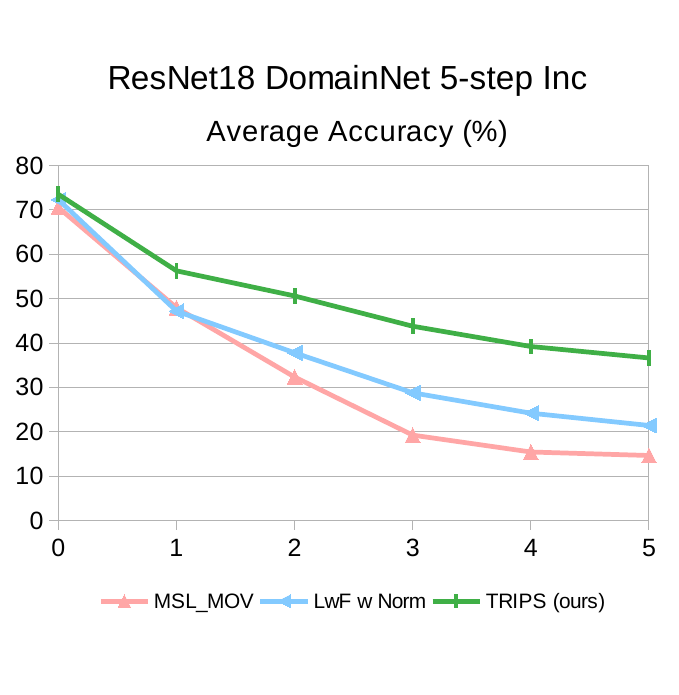}
    \label{fig: ResNet18_DomainNet_5_step_acc}
	\end{minipage}
    \begin{minipage}[t]{0.47\linewidth}
    \includegraphics[trim={9cm 2cm 7.5cm 2cm},clip,width=\linewidth]{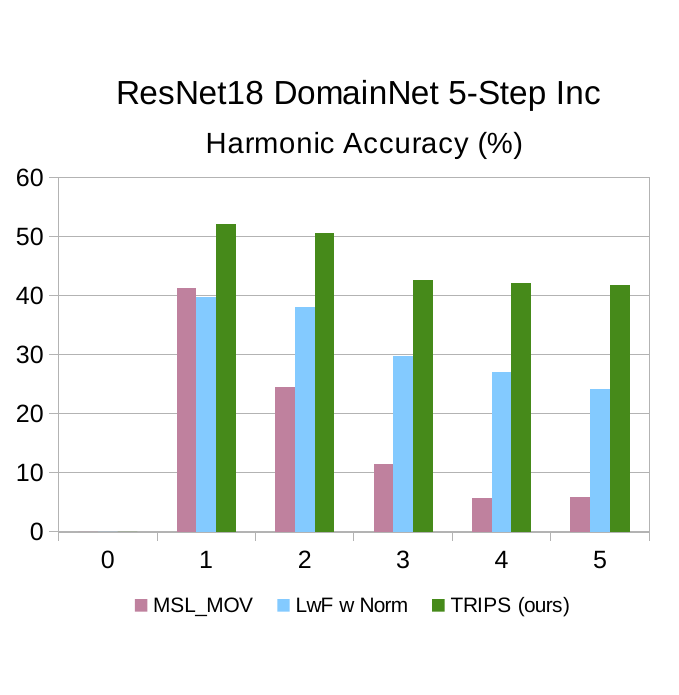}
    \label{fig: ResNet18_DomainNet_5_step_harm_acc}
    \end{minipage}
    \vspace{-0.8cm}
    \caption{Performance under {\em the five-step incremental settings} (20 new classes added at a time) on the DomainNet dataset based on ResNet-18.}
    \label{fig: ResNet18_DomainNet}
\end{figure}

\subsection{Backbone Analysis}
\label{sub-sec: Different Backbones}
To validate that the benefits of the proposed method are independent of the used backbone network,
we  perform experiments based on other backbones instead of ResNet-34.
Figures \ref{fig: ResNet18_OfficeHome} and \ref{fig: ResNet18_DomainNet} show the comparison with other methods on the OfficeHome \citep{venkateswara2017deep} and DomainNet \citep{saito2019semi} datasets using ResNet-18 as the backbone network.
In both settings, our TRIPS method always achieves the best performance.

\subsection{Training Time Analysis}
\label{sub-sec: Training Time Analysis}
We run all our experiments on 1x NVIDIA v100 GPU with 16GB RAM. 
We use a batch size of 32 for each domain set, and three training domains are utilized in each batch. Thus, the total batch size is 96. 
When data augmentation is utilized, due to memory limitations, we reduce the batch size from 32 to 24 for each domain set, thus the total batch size is reduced from 96 to 72. 
For each incremental session, the training time is between 5 and 7 hours. 
When data augmentation is utilized, the training data within each batch is enlarged from 72 to 288, which causes the training time to increase to be between 10 and 15 hours long per incremental session. 
Within each batch, an equal number of pseudo-samples from old classes will be generated from the multivariate Normal distributions (this step do not rely on a feature extractor  minimizing the impact on runtime). These samples of old classes are used along the samples from the new classes.



\section{Conclusions}
\label{sec:concl}
In this paper, we have explored the challenging DGCIL task and proposed our TRIPS method. 
To improve the generalization of the model, triplet loss helps obtain semantic information while eliminating domain information. 
With no old exemplars preserved, we sample pseudo-features of old classes from updated prototype representations (multivariate Normal distributions). 
These pseudo-features not only help preserve old class knowledge but also support the TRIPS loss to maintain inter-class boundaries, thus boosting the performance. Importantly, these prototype representations are domain-agnostic as our loss enforces domain confusion.
Substantial improvements in benchmarks demonstrate the efficiency of our method.

\section*{Acknowledgements}
This work was funded by CSIRO's Reinvent Science and CSIRO's Data61 Science Digital. 

\balance{}

\bibliographystyle{model2-names}
\bibliography{refs}

\clearpage
\appendix

\section*{Supplementary Material}
\setcounter{section}{0}

\section{Ablation Study}
To validate the effectiveness of each part of our method, we perform the ablation study on OfficeHome \citep{venkateswara2017deep}.
The experimental results are shown in Table \ref{tab: Ablation_OfficeHome} and \ref{tab: sub_Ablation_OfficeHome}.
Firstly, to improve the generalization of the model, the triplet loss is applied to extract semantic information while suppressing domain information.
Observing the experimental results, the benefit of triplet loss is clear.
Then, we update old class prototypes and sample pseudo-old class features from the updated prototypes. 
With the help of the multivariate sampling strategy, the performance is further boosted. 

\begin{table*}
\setlength\tabcolsep{3pt}
\centering
\footnotesize{
	\caption{Ablations on OfficeHome under \emph{the 5-step incremental setting} (10 new classes added at a time). Results use format $a\: (b, c, d, e)$ where $a$ is (top) the average and (bottom) harmonic accuracy over all four test domains, whereas $(b, c, d, e)$ represents (top) the average and (bottom) harmonic accuracy per test domain: art, clipart, product, real.}
    \vspace{-0.2cm}
	\begin{tabular}{c c c c | c c c c c c}
		\hline
		\multirow{2}*{KD} & \multirow{2}*{triplet} & \multirow{2}*{\tabincell{c}{proto \\ shift}} & \multirow{2}*{\tabincell{c}{pseudo \\ sampling}} & 0 (base) & 1 & 2 & 3 & 4 & 5 \\
        \cline{5-10}
        & & & & \multicolumn{6}{c}{Domain-Average ( Art, Clipart, Product, Real)} \\
        \hline
        & & & & \multicolumn{6}{c}{average accuracy (\% $\uparrow$)} \\
        \hline
        \checkmark &  &  &  & 81.29 & 50.89 & 41.37 & 34.52 (27.64, 32.96, 37.60, 39.88) & 31.67 (24.97, 30.04, 33,36, 38.31) & 27.94 (21.65, 26.54, 28.37, 35.21) \\
        \checkmark  & \checkmark &  &  & \textbf{82.89} & 60.85 & 52.97 & 47.62 (36.95, 40.89, 53.56, 59.08) & 44.81 (34.27, 37.98, 49.69, 57.32) & 40.18 (31.32, 33.47, 44.71, 51.19) \\
        \checkmark  & \checkmark & \checkmark &  & \textbf{82.89} & 60.73 & 53.38 & 48.27 (37.01, 41.98, 55.20, 58.90) & 44.68 (34.21, 38.42, 49.94, 56.14) & 39.92 (30.58, 33.84, 44.38, 50.86) \\
        \rowcolor{LightCyan} \checkmark  & \checkmark & \checkmark & \checkmark & \textbf{82.89} & \textbf{64.08} & \textbf{55.18} & \textbf{49.25 (37.76, 42.74, 55.45, 61.03)} & \textbf{45.58 (34.71, 38.45, 51.06, 58.12)} & \textbf{40.74 (31.39, 33.97, 45.49, 52.13)} \\
        \hline
        & & & & \multicolumn{6}{c}{harmonic accuracy (\% $\uparrow$)} \\
        \hline
        \checkmark &  &  &  & - & 44.45 & 41.58 & 37.49 (30.07, 35.84, 40.84, 43.21) & 35.52 (27.76, 33.57, 37.55, 43.19) & 32.22 (24.81, 30.46, 32.83, 40.79) \\
        \checkmark & \checkmark &  &  & - & 58.06 & 54.35 & 48.83 (38.95, 42.26, 54.28, 59.84) & 46.42 (37.18, 38.61, 50.81, 59.09) & 40.30 (32.65, 32.96, 44.41, 51.16) \\
        \checkmark & \checkmark & \checkmark &  & - & 58.09 & 55.51 & 49.63 (39.24, 42.90, \textbf{56.59}, 59.80) & 46.50 (37.00, \textbf{38.92}, 51.58, 58.52) & 39.95 (31.68, 33.96, 44.54, 49.63) \\
        \rowcolor{LightCyan} \checkmark  & \checkmark & \checkmark & \checkmark & - & \textbf{61.45 } & \textbf{57.11} & \textbf{49.76 (39.26, 43.62,} 55.96, \textbf{60.19)} & \textbf{47.22 (37.50,} 38.58, \textbf{52.73, 60.06)} & \textbf{41.51 (33.56, 34.55, 45.54, 52.37)} \\
        \hline
	\end{tabular}	
	\label{tab: sub_Ablation_OfficeHome}
}
\end{table*}

\end{document}